\documentclass[10pt,twocolumn,letterpaper]{article}

\usepackage{cvpr}              %

\usepackage[dvipsnames]{xcolor}

\usepackage{times}
\usepackage{epsfig}
\usepackage{graphicx}
\usepackage{amsmath}
\usepackage{amssymb}
\usepackage{bbm}
\usepackage{booktabs}
\usepackage{bm}
\usepackage{multirow}
\usepackage{microtype}
\usepackage[T1]{fontenc}
\usepackage{soul}

\usepackage{xcolor}
\usepackage{listings}
\usepackage{paralist}
\usepackage[toc,page]{appendix}

\definecolor{cvprblue}{rgb}{0.21,0.49,0.74}
\usepackage[pagebackref,breaklinks,colorlinks,citecolor=cvprblue]{hyperref}
\usepackage{tabularx}
\usepackage{color,soul}
\usepackage{multirow}
\usepackage{makecell}
\usepackage{xcolor,colortbl}
\usepackage{wrapfig}
\usepackage[flushleft]{threeparttable}
\usepackage{fixltx2e}
\usepackage{hyperref}
\usepackage{xspace}
\usepackage{graphicx}
\usepackage{placeins}
\usepackage{caption}
\usepackage[export]{adjustbox}[2011/08/13]  %
\definecolor{tablegreen}{rgb}{0.9,0.99,0.9}
\def\paperID{7674} %
\def\confName{CVPR}
\def\confYear{2024}

\title{Iterated Learning Improves Compositionality in Large Vision-Language Models}

\author{%
Chenhao Zheng$^{2}$, \
Jieyu Zhang$^{1}$, \
Aniruddha Kembhavi$^{3}$, \
Ranjay Krishna$^{1,3}$\\ 
$^1$University of Washington,
$^2$University of Michigan,
$^3$Allen Institute for Artificial Intelligence\\
\texttt{neymar@umich.edu, \{jieyuz2,ranjay\}@cs.washington.edu, anik@allenai.org}
}

\newcommand{\modify}[1]{#1}

\begin{document}
\maketitle

\begin{figure*}[h]
  \centering
    \includegraphics[width=\textwidth]{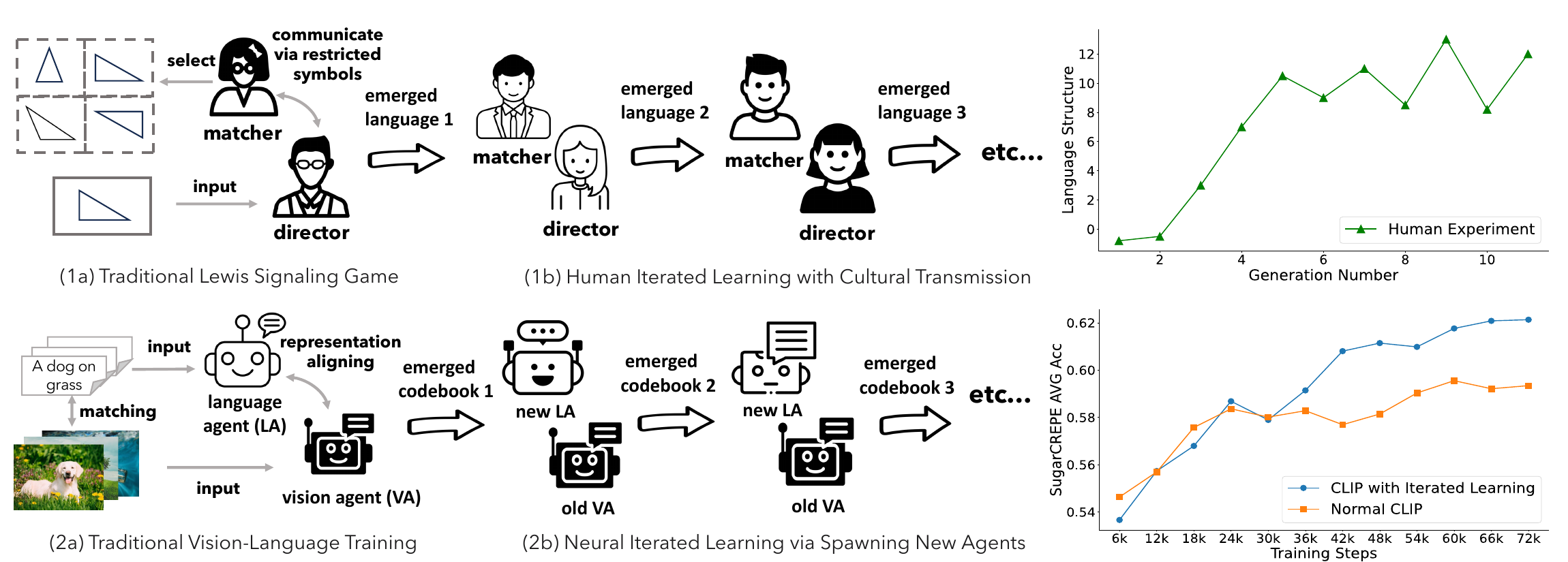}
    \caption{(1). From studying the language that emerged from Lewis Signaling Game, evolutionary linguistics found that iterated learning with cultural transmission leads to language compositionality. (2). We interpret vision-language model training as Lewis Signaling Game between neural agents, and discovered iterated learning can also improve the compositionality of vision-language model's representation  } 
    \label{fig:teaser}
\end{figure*}

\begin{abstract}

A fundamental characteristic common to both human vision and natural language is their compositional nature. Yet, despite the performance gains contributed by large vision and language pretraining, recent investigations find that most—if not all—our state-of-the-art vision-language models struggle at compositionality.
They are unable to distinguish between images of ``a girl in white facing a man in black'' and ``a girl in black facing a man in white''.
Moreover, prior work suggests that compositionality doesn't arise with scale: larger model sizes or training data don't help.
This paper develops a new iterated training algorithm that incentivizes compositionality.
We draw on decades of cognitive science research that identifies \textit{cultural transmission}—the need to teach a new generation—as a necessary inductive prior that incentivizes humans to develop compositional languages. 
Specifically, we reframe vision-language contrastive learning as the Lewis Signaling Game between a vision agent and a language agent,  
and operationalize cultural transmission by iteratively resetting one of the agent's weights during training.
After every iteration, this training paradigm induces representations that become ``easier to learn'', a property of compositional languages: e.g. our model trained on CC3M and CC12M improves standard CLIP by 4.7\%, 4.0\% respectfully in the SugarCrepe benchmark.

\end{abstract}

\section{Introduction}
\label{sec:intro}

Scholars across disciples herald \emph{compositionality} as a fundamental presupposition characterizing both human perception and linguistic processing~\cite{cresswell1973logics,fodor1988connectionism}.
Through compositional reasoning, humans can comprehend the photos they take and describe those images by composing words together~\cite{janssen1997compositionality,hupkes2020compositionality,bottou2014machine,chomsky1965some}.
For instance, compositionality allows people to differentiate between a photo of ``a gold colored dog facing a person wearing black'' and ``a black colored dog facing a person wearing gold''.
Given its importance, research in both computer vision and natural language processing has sought to develop models that can similarly comprehend scenes and express them through compositional language~\cite{krishna2017visual,ji2020action,lu2016visual,GrundeMcLaughlin2021AGQA}.

Yet, a series of recent evaluation benchmarks conclude that state-of-the-art vision-language models exhibit little to no compositionality~\cite{ma2022crepe,hsieh2023sugarcrepe,thrush2022winoground,zhao2022vl,yuksekgonul2023when,ray2023cola}.
In fact, in many specific evaluation conditions, models perform almost close to random chance. 
Even models such as CLIP~\cite{radford2021learning}, which has become the backbone for many vision tasks, exhibit little compositionality.
More striking are the experiments that suggest that compositionality doesn't emerge with scale, \ie vision models do not become more compositional with increasing model size or training data~\cite{ma2022crepe,hsieh2023sugarcrepe}.
Similar experiments in natural language processing find that large language models also struggle with compositionality~\cite{dziri2023faith,berglund2023reversal}.

Meanwhile, Cognitive Scientists have spent the last two decades studying the emergence of compositionality in human language. The results seem to indicate that the primary inductive prior that leads to language compositionality is \textit{cultural transmission}: a phenomenon where an older generation transmits their language to a new generation~\cite{townsend2018compositionality,verhoef2016iconicity,brighton2006understanding,carr2017cultural}. They hypothesize that this need to teach our offsprings our language creates a natural preference towards languages that are easier to learn. A compositional language, which necessitates learning only a limited number of symbols to express infinite concepts, is therefore preferred to ones with unique symbol-to-concept bindings.

To demonstrate this hypothesis, scientists study the language that emerged from the ``Lewis Signaling Game''. Lewis Signaling Game~\cite{lewis2008convention} is a theoretical framework 
where two people communicate with one another to solve the ``object reference'' problem (Fig.~\ref{fig:teaser}(1a)). Their communication channel is restricted to symbols, which do not represent any known language, forcing participants to develop a new shared language to communicate. They simulate cultural transmission by replacing human participants across ``generations'', and observe how new combinations of participants modify their language (Fig.~\ref{fig:teaser}(1b)). 
They verify that over multiple generations, the emergent language becomes more compositional~\cite{kirby2001spontaneous,cogswell2019emergence,guo2019emergence,li2019ease,cornish2017sequence,verhoef2016iconicity}.

In this paper, we operationalize cultural transmission as an iterated learning (IL) algorithm for vision-language models.
Consider the popular CLIP model; it is trained to learn representations through an interplay between vision and language representations~\cite{radford2021learning}. At a high level, its contrastive learning objective trains image representations that can retrieve their corresponding textual description from a set of distractors, and vice versa.
We reframe this objective through the lens of the Lewis Signaling Game (Fig.~\ref{fig:teaser}(2a)). 
Similar to the cognitive science studies that involve two human participants, vision-language training can be viewed as a game between two model participants: a vision agent and a language agent attempting to learn a shared representation.
With this framing in mind, we apply cultural transmission by periodically spawning a new language agent to replace the old one (Fig.~\ref{fig:teaser}(2b)).
Intuitively, the need to re-train a new language agent is akin to ``teaching a new generation'' and should similarly encourage the vision agent to produce representations that are easier to learn. 
We also create the notion of ``shared and limited communication symbols'' by learning a shared codebook as the basis for representations that both agents can use.

Our experiments demonstrate that our algorithm does in fact induce easy-to-learn representation, improving compositionality in vision-language models. For example, our model trained in CC3M improves standard CLIP by 4.7\% in SugarCrepe~\cite{hsieh2023sugarcrepe} and by 3.8\% in CREPE~\cite{ma2022crepe}, both benchmarks are specially designed for testing compositionality for vision-language models.
Notably, our model exhibits better compositionality than existing compositional methods, such as NegCLIP~\cite{yuksekgonul2022and}. Our model does not require extra training time despite periodically resetting model weights, and does not harm the CLIP's recognition performance.
We also demonstrate the easy-to-learn property in our representation in experiments and find that the emerged codebook contains interpretable concepts.

\section{Related Works}

Our work is inspired by cognitive science literature and the related works span various areas, including large vision-language models, the emergence of language, and interacting neural agents. 

\noindent\textbf{Compositionality in vision-language models.} With the popularity of the CLIP model~\cite{radford2021learning}, contrastive learning has become the de-facto way of aligning representations for different modalities~\cite{rombach2022high,girdhar2023imagebind,li2022blip,huang2023language,yu2022coca, yang2024unitouch, zheng2023exif}. However, despite their remarkable ability in zero-shot recognition \cite{radford2021learning}, their features exhibit little compositionality~\cite{thrush2022winoground,zhao2022vl,yuksekgonul2023when,ma2022crepe,ray2023cola,hsieh2023sugarcrepe}. 
For example, all models struggle to identify the captions with correct word order~\cite{yuksekgonul2022and}, compose concepts together to express compositional concepts \cite{ma2022crepe}, and compose attributes and relations~\cite{zhao2022vl, yuksekgonul2022and, ray2023cola, hendricks2021probing}. Attempts have been made to enhance CLIP's compositionality, including hard-negative mining~\cite{yuksekgonul2022and}, cleaning the data~\cite{li2022blip}, and using novel representation formats~\cite{chen2023revisiting,zhong2021learning, liu2021cross}. However, the recently proposed SugarCrepe benchmark~\cite{hsieh2023sugarcrepe} finds that their improvements are overestimated, calling for a more effective method.

\noindent\textbf{Iterated learning and cultural transmission.} Human language is, for the most part, compositional. Evolutionary linguists have spent decades studying the origin of compositionality of human language~\cite{pinker1990natural,perfors2002simulated}. One important factor appears to be the need to transmit the language across multiple generations~\cite{kirby2014iterated}, formulated by Kirby~\cite{kirby2008cumulative, kirby2014iterated, kirby2001spontaneous} as a framework called iterated learning. Extensive simulations~\cite{christiansen2003language, cangelosi2012simulating, smith2003iterated} and human experiments~\cite{kirby2008cumulative, cornish2017sequence,silvey2015word,kirby2014iterated} demonstrate its ability to incentivize the emergence of compositional structure in their language, in small-scale and quantized environments. Newer experiments in open and continuous environments also conclude similar findings~\cite{xu2013cultural,carr2017cultural,silvey2013communication}, although they observe a large amount of randomness across experiments~\cite{carr2017cultural}.

\noindent\textbf{Emergence of linguistic structure in neural agents.}
Collaborative AI agent systems have been the subject of much research, in which neural agents communicate to learn a language while accomplishing goals~\cite{kottur2017natural,lazaridou2018emergence, dessi2021interpretable, ren2020compositional, lazaridou2016multi}. Most approaches learn a discrete communication protocol while playing the Lewis Signaling Game ~\cite{kottur2017natural, li2019ease, cogswell2019emergence, ren2020compositional, lazaridou2016multi}.  
Researchers find the language developed by agents, if compositional, shows enhanced systematic generalization capabilities \cite{kottur2017natural, ren2020compositional, chaabouni2020compositionality}. However, compositionality does not occur naturally \cite{chaabouni2020compositionality, kottur2017natural} and is not tied to generalization pressure \cite{kharitonov2020emergent}. Some works introduced neural iterated learning frameworks \cite{cogswell2019emergence, ren2020compositional, li2019ease, rajeswar2022multi, vani2021iterated, ren2024improving}. Using topographic similarity as a measurement, they found that emergent language is more compositional \cite{ren2020compositional, cogswell2019emergence, li2019ease}.  Some works also show the resulting compositional language is easier to learn ~\cite{rita2022role, li2019ease}, corresponding to the finding in cognitive science \cite{kirby2014iterated}.  However, the experiments are limited to small domains with easily-categorizable inputs like simple cubics or balls \cite{kottur2017natural, ren2020compositional, li2019ease, lazaridou2016multi}. The message structures and network architecture are also simple, raising the question of scalability. Our work is similar in the idea of using iterated learning to boost compositionality. However, our model observes large-scale real-world data that are not easily categorizable and uses contrastive learning as opposed to reinforcement learning used by most methods.

\section{Method}
\label{sec:method}

\begin{figure*}[t]
  \centering
    \includegraphics[width=\textwidth]{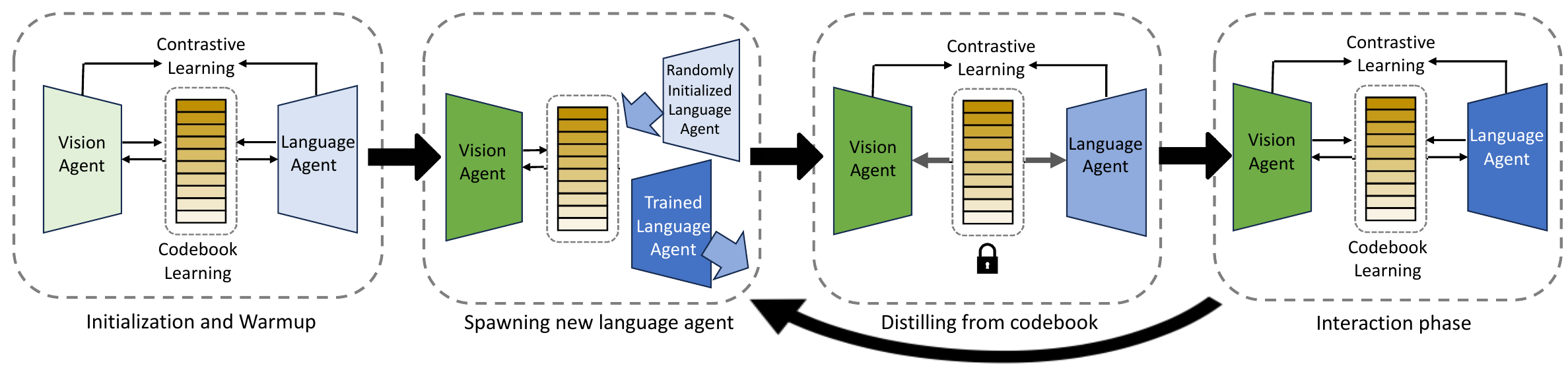}
    \caption{Our iterated learning algorithm is built on CLIP augmented with a shared codebook. The algorithm consists of a warmup stage and three iterated phases that cycle until the end of training.  In each cycle, we 1) spawn a new language agent to replace the old one. 2) frozen codebook weight for a certain number of steps. 3) let agents interact under standard vision-language contrastive learning.
    }
    \label{fig:method}
\end{figure*}

We design an iterated learning algorithm to improve the compositionality of vision-language models.
To do so, we draw an analogy between the process of vision-language contrastive learning and the procedure of Lewis Signal Game \cite{lazaridou2018emergence}, and build our method upon CLIP's training objective~\cite{radford2021learning}.
We first reframe CLIP as Lewis Signaling Game (Sec.~\ref{subsec: reframe}); then we introduce the shared codebook module that bottlenecks each modalities'  representations (Sec.~\ref{subsec:codebook}); finally, we describe our iterated learning algorithm (Sec.~\ref{subsec:iterated}).

\subsection{Reframing vision-language contrastive learning as a Lewis Signaling Game}
\label{subsec: reframe}
In the traditional Lewis Signaling Game, two people communicate through restricted symbols to solve a referential task. In the task, one person called the ``director'' observes an input stimuli (usually a picture of abstract shapes) and needs to choose a sequence of symbols from a limited vocabulary to send over to the second person, the ``matcher''. The matcher sees only the symbols and a set of observations, from which they must identify the one seen by the director.
The evolving conversation patterns across time are treated as emergent language. 
This game setup is similar to the contrastive learning procedure popular today in vision-language training, where a vision agent and a language agent observe modality-specific inputs and need to communicate together to identify the matching image-text pairs from distractors.

More formally, during the training process, two agents observe their distinctive inputs (images $\{u_i\}_{i=1}^N$ for the vision agent and texts $\{v_i\}_{i=1}^N$ for the language agent). They encode the inputs to representations $(f_\theta(u_i), g_\phi(v_i))$, which serves as the cross-agent communication messages. The communication objective is that, given $N$ images and $N$ pieces of text, the corresponding image-text pairs should be successfully matched, implemented using the contrastive objective:
\begin{equation}
  \mathcal{L} = -\sum_{i=1}^N \log \frac{\exp{(f_\theta(u_i) \cdot g_\phi(v_i) / \tau)}}{\sum_{j=1}^N \exp(f_\theta(u_i) \cdot g_\phi(v_j)/\tau)},
  \label{eq:contrast}
\end{equation}
where $\tau$ is a small constant. The final aligned representation $(f_\theta(u_i), g_\phi(v_i))$ can be viewed as \textit{the shared language} emergent between the two agents.

\subsection{Shared codebook for a regulated representation}
\label{subsec:codebook}
One of the key designs of the Lewis Signaling Game is the limited vocabulary that participants can use, while in vision-language contrastive learning the agents don't have any regulation on the representations they use to communicate. 
Therefore, to follow the Lewis Signaling Game, we employ a learnable codebook as the basis of representations shared by agents to regulate their representation space.
\modify{In particular, the codebook is composed of a finite number of codes, representing shared and limited ``vocabularies`` in the learned language. The final layer of vision and language encoders sparsely combines the codebook to produce the final representation, representing the learnable ``vocabulary composition rule`` }

Let $\{c_i\}_{i=1}^C$ be a codebook, where $C$ is the predefined number of codes. We use the Transformer architecture for both agents. Thus, given an input image $u$, the vision agent $f$ extracts patch embeddings $p_j$ for each patch $j$ from the transformer's last-layer activations. We define the similarity score between code $c_i$ and the image $u$ as the maximum cosine value between the code and patch features: 
\begin{equation}
r_i^u=\max _j<f_{p_j}, c_i>
 \label{eq:relevant}
\end{equation}
This codebook architecture is derived from recent work using codebooks for vision-language training~\cite{chen2023revisiting}.
Following \cite{chen2023revisiting}, we normalize $r_i^u$ using Sparsemax function \cite{martins2016softmax}, which generates a \textit{sparse} similarity score $w_i^v$ for each code.  The output representation for the input image $u$ is the linear combination of codes $c_i$, with $w_i^v$ being multiplied as weights:
\begin{equation}
f(u)=\sum_i^C w_i^v \cdot c_i
\end{equation}
The procedure for obtaining text representation $g(v)$ using the language agent $g$ is defined analogously. Instead of patch embeddings for the vision agent, here, we use the language input's token embeddings.

\subsection{Iterated learning in training} 
\label{subsec:iterated}

Our iterated learning algorithm consists of a warmup stage, followed by $K$ training cycles; each mirrors the concept of 'generations' in cultural transmission theory and consists of three phases: spawning a new language agent, distillation from the codebook, and an interaction phase. We visualize the algorithm in Fig.~\ref{fig:method}.

\noindent\textbf{Initialization and warmup stage.} 
The beginning of our training algorithm is similar to CLIP's algorithm. We randomly initialize the parameters of both the agents and let them train for $T_{warm}$ number of iterations. 

\noindent\textbf{Spawning a new language agent.}
This stage simulates introducing a new participant that replaces an older one, representing a new generation in cultural transmission.
While studies in cognitive science~\cite{brighton2006understanding,carr2017cultural} replace both participants over multiple generations, our ablation study indicates that replacing both is unnecessary; it even increases the training time required to achieve the same level of compositionality. 
By contrast, we replace only the language agent between generations by reinitializing it with random parameters.
Although outside the scope of this paper, we hypothesize that resetting just the language agent works better empirically because the vision agent needs to simultaneously learn lower-level visual features and also associate them with high-level concepts while the language agent only needs to learn to extract high-level concepts from text.

\noindent\textbf{Distilling from the codebook.} 
Serving as the basis for both agent's representations, the quality of the learned codebook is essential.
We find that introducing a new agent, with its randomly initialized and under-trained weights, leads to large changes to the codebook gradients, causing instability in training.
We, therefore, add a distillation stage to ensure that the codebook evolves smoothly across generations.
The older language agent is distilled into the new language agent for $T_{distill}$ iterations~\cite{hinton2015distilling}.
We temporarily freeze the codebook's weights during this phase. This allows the new agent to adapt to the existing codebook structure without introducing disruptive changes.
Unlike traditional distillation, this phase does not train till convergence. After $T_{distill}$ steps, we switch to the interaction phase.

\begin{table*}[tbp]
\centering

\scalebox{0.89}{

\setlength{\tabcolsep}{5pt}
\begin{tabular}{l l  ccccc ccc cc c c c} 
    \toprule
    \multirow{2.5}{*}{\textbf{Dataset}} & 
   \multirow{2.5}{*}{\textbf{Method}} & 
    \multicolumn{2}{c}{\textbf{CREPE-systematicity}} & 
    \multicolumn{3}{c}{\textbf{CREPE-productivity}} & 
    \multicolumn{3}{c}{\textbf{SugarCrepe}} &
    \textbf{Cola} &
    \multicolumn{1}{c}{\textbf{\small Winoground}} & 
    \multirow{2.5}{*}{\textbf{\small Mean}} \\
    \cmidrule(lr){3-4}
    \cmidrule(lr){5-7}
    \cmidrule(lr){8-10}
    \cmidrule(lr){11-11}
    \cmidrule(lr){12-12}

   & & {\small atom} & {\small compound} & {\small replace} & {\small swap}& {\small negate}& {\small add} & {\small replace} & {\small swap} &  {\small Txt2Img} & {\small Txt2Img} \\ 
    \midrule

    \multirow{4}{*}{CC3M} & CLIP~\cite{radford2021learning} &  28.1 & 38.4 & 9.8 & 18.1 & 4.0 & 61.9 & 64.3 & 52.9 & 17.6 &  8.1 & 28.3    \\
    & Codebook-CLIP~\cite{chen2023revisiting} & 28.8 & 40.3 & 10.9 & 19.2 & 3.5 & 65.9 & 64.8 & 54.9 & 15.7 & 8.8 & 31.2   \\
    & NegCLIP\textsuperscript{*} ~\cite{yuksekgonul2022and}     & 29.5 & 41.8 & 11.6 & \textbf{33.3} & \textbf{5.8} & 59.3 & 59.2 & \textbf{60.1} & 16.5 & 11.8 & 32.8 \\
    
    & \textbf{IL-CLIP (Ours)} & \textbf{33.2} &  \textbf{47.7} & \textbf{14.6} & 22.3 & 5.3 & \textbf{66.1} & \textbf{67.0} & 54.5 & \textbf{20.0} & \textbf{13.3} & \textbf{34.4} \\

    \midrule

    \multirow{4}{*}{CC12M} & CLIP~\cite{radford2021learning} & 35.0 & 42.7 & 12.3 & 19.5 & 14.6 & 67.5 & 70.0 & 60.2 & 21.5 & 7.2 & 34.9\\
    & Codebook-CLIP~\cite{chen2023revisiting} & 35.6 & 43.9 & 14.4 & 22.0 & 12.8 & 71.3 & 71.1 & 59.5 & 20.8 & 9.5 & 36.1  \\
    & NegCLIP\textbf{\textsuperscript{*}} ~\cite{yuksekgonul2022and} & \textbf{36.6} & 45.2 & 14.9 & \textbf{35.8} & \textbf{15.2} & 65.0 & 70.2 & \textbf{67.2} & \textbf{22.7} & 7.3 & \textbf{38.0} \\
    & \textbf{IL-CLIP (Ours)} & \textbf{36.6} & \textbf{47.5} & \textbf{17.9} & 23.9 & 14.8 & \textbf{73.8} & \textbf{73.0} & 62.9 & 20.2 & \textbf{10.1} & \textbf{38.0} \\
    
    \bottomrule
\end{tabular}
}
\vspace{-2mm}
\caption{\textbf{Evaluation on compositionality benchmarks.} We do image-to-text retrieval on CREPE systematicity-CC12M split, CREPE productivity split, and SugarCrepe \cite{ma2022crepe, hsieh2023sugarcrepe}. We do text-to-image retrieval on Cola and Winoground \cite{diwan2022winoground,ray2023cola}. We report the retrieval R@1 scores. IL-CLIP notably improves CLIP's compositionality, and exhibits better performance than NegCLIP in most datasets. \textbf{(*)} Note that NegCLIP directly trains on the text negatives close to ``swap'' objectives, and therefore obtains unusually high scores for that split.
 }\vspace{-2mm} %
\label{tab:comp_result}
\end{table*}
\setlength{\tabcolsep}{6pt}

\begin{table*}[h]

    \centering
\scalebox{0.71}{
    \begin{tabular}{l l c c c c c c c c c c c c c c c c c c c} 
        \toprule
        \textbf{Dataset}  &  
        \textbf{Method}  &  
        \rotatebox[origin=lb]{90}{\textbf{ImageNet1k}} &  
        \rotatebox[origin=lb]{90}{\textbf{CIFAR-100}} &  
        \rotatebox[origin=lb]{90}{\textbf{CIFAR-10}} &  
        \rotatebox[origin=lb]{90}{\textbf{STL-10}} &  
        \rotatebox[origin=lb]{90}{\textbf{VOC2007}} &  
        \rotatebox[origin=lb]{90}{\textbf{Caltech101}} &  
        \rotatebox[origin=lb]{90}{\textbf{SUN397}} &  
        \rotatebox[origin=lb]{90}{\textbf{Pets}} &  
        \rotatebox[origin=lb]{90}{\textbf{Flowers102}} &  
        \rotatebox[origin=lb]{90}{\textbf{Food101}} &  
        \rotatebox[origin=lb]{90}{\textbf{ObjectNet}} &  
        \rotatebox[origin=lb]{90}{\textbf{CLEVR}} &  
        \rotatebox[origin=lb]{90}{\textbf{Smallnorb}} &  
        \rotatebox[origin=lb]{90}{\textbf{Resisc45}} & 
        \rotatebox[origin=lb]{90}{\textbf{DMLAB}} & 
        \rotatebox[origin=lb]{90}{\textbf{ImageNet-A}} & 
        \rotatebox[origin=lb]{90}{\textbf{ImageNet-R}} & 
        \rotatebox[origin=lb]{90}{\textbf{IN-sketch}} & 
        \rotatebox[origin=lb]{90}{\textbf{Mean}}
        \\ 
        
        \midrule %
        & CLIP \cite{radford2021learning} &  
        13.7 & 18.6 & 43.5 & 80.7 & 44.3 & 60.1 & 28.6 & 8.9 & 9.1 & 8.3 & 8.0 & \textbf{19.5} & 5.2 & 12.6 & 11.7 & 3.0 & 17.7 & 7.2 & 21.7     \\
        \multirow{2}{*}{CC3M} & Codebook-CLIP \cite{chen2023revisiting} & 
        \textbf{14.8} & \textbf{22.0} & \textbf{49.8} & 85.4 & \textbf{48.3} & 60.8 & 30.4 & 8.8 & 8.5 & \textbf{10.5} & \textbf{9.1} & 16.7 & 4.8 & \textbf{19.8} & 17.5 & \textbf{3.7} & \textbf{20.1} & \textbf{8.2} & \textbf{24.4}  \\
        & NegCLIP \cite{yuksekgonul2022and} & 
        11.8 & 19.6 & 44.0 & 78.2 & 44.6 & 52.1 & 25.8 & 9.1 & 8.6 & 6.6 & 6.9 & 15.1 & 6.2 & 13.8 & 11.9 & 2.4 & 15.8 & 5.1 & 21.0 \\
        & \textbf{IL-CLIP (Ours)} & 
        14.2 & 20.9 & 48.6 & \textbf{87.7} & \textbf{48.3} & \textbf{61.1} & \textbf{32.8} & \textbf{10.0} & \textbf{9.2} & 9.1 & 8.4 & 15.8 & \textbf{5.5} & 15.6 & \textbf{18.7} & 2.9 & 18.8 & 6.5 & 24.2\\

        \midrule %
        & CLIP \cite{radford2021learning} & 
        31.4 & 30.9 & 60.1 & 89.3 & 53.3 & 72.5 & 41.0 & 49.6 & 21.1 & 31.5 & 17.8 & 20.0 & 11.7 & 26.5 & 13.6 & 4.4 & 44.2 & 24.0 & 35.7  \\

        \multirow{2}{*}{CC12M} & Codebook-CLIP \cite{chen2023revisiting} & 
        \textbf{34.2} & \textbf{39.6} & \textbf{68.1} & 90.3 & 55.5 & 75.4 & 45.8 & \textbf{53.9} & \textbf{24.8} & \textbf{32.3} & 20.4 & \textbf{24.0} & \textbf{15.5} & 27.6 & 11.7 & 5.2 & 48.8 & \textbf{26.9} & \textbf{38.8}\\
        
        & NegCLIP \cite{yuksekgonul2022and} & 
        28.9 & 27.1 & 55.4 & 89.7 & 54.1 & 72.8 & 42.6 & 44.6 & 22.3 & 30.2 & 17.8 & 17.5 & 10.5 & 26.2 & \textbf{15.9} & 4.1 & 39.6 & 22.0 & 34.5 \\
        
        & \textbf{IL-CLIP (Ours)} & 32.8 & 32.5 & 61.6 & \textbf{94.1} & \textbf{60.0} & \textbf{76.9} & \textbf{49.7} & 51.6 & 21.4 & 31.8 & \textbf{22.7} & 20.6 & 12.9 & \textbf{27.7} & 15.3 & \textbf{7.2} & \textbf{49.0} & 25.6 & 38.5 \\
        \bottomrule
    \end{tabular}
    }
    \caption{\textbf{Evaluation of zero-shot image classification on 18 commonly used public datasets.} Scores are reported in terms of top-1 accuracy. Using a shared codebook (Codebook-CLIP) boosts standard CLIP's classification performance, and adding our iterated learning paradigm on top of Codebook-CLIP (IL-CLIP) does not sacrifice the overall performance. 
    }
    \label{tab:classification}
    \vspace{-5pt}
\end{table*}

\noindent\textbf{Interaction phase.} After distillation, we unfreeze the codebook and train the model normally following the standard vision-language contrastive learning paradigm~\cite{radford2021learning}. 
By letting agents interact freely, we expect their representations to begin aligning with one another again. We also limit the duration of this interacting phase to be $T_{interact}$ step, ensuring a learning bottleneck such that the education process from the old vision agent to the new language agent is incomplete and biased.

After the interacting stage, the current generation of agents is considered finished and the next generation begins. We repeat the above three phases until the end of training. During the last phase, we extend the interaction phase to allow training till convergence.

\noindent\textbf{Understanding our algorithm.}
\textit{From a cognitive science perspective}, the ``knowledge gap'' between old and new agents creates an implicit ``teaching'' scenario, where the vision agent interacts with the newly initialized language agent to realign both their cross-modal representations. This pressure to teach, as posited in cultural transmission theory, encourages the developed representations to be easier for subsequent agents to learn, potentially leading to better compositionally. \modify{ We empirically demonstrate this "easy-to-learn" property at Sec.~\ref{subsec: evolution}.}

\textit{From a machine learning perspective}, theory and results suggest that self-distillation performs label smoothening~\cite{zhang2020self} and smoothness regularization in the function space~\cite{mobahi2020self}.  It reinforces the optimization bias of neural networks for smooth
solutions~\cite{rahaman2019spectral}. In other words, distillation with early stopping—like the one we are doing—makes the new generation a smoother low-frequency approximation of the older generation.
During the interaction phase, the vision agent adjusts its parameters to align better with this newer, smoother language agent. Since smoother functions are characterized by a smaller Lipschitz constant, they are easier to learn; therefore, every iteration should lead to easier-to-learn functions. Since compositional languages are easier to learn, every cycle possibly makes the representations more compositional.
We observe this phenomenon empirically in our experiments.  \modify{At Sec.~\ref{subsec: evolution}, we show through experiment that the upper bound of Lipschitz constant indeed decreases over time.}

\section{Experiment}
\label{sec:experiment}
Our experiments evaluate both the compositionality (Sec. ~\ref{subset:compositionality}) and recognition capability (Sec.~\ref{subset:recognition}) of the trained representation. In Sec.~\ref{subsec: evolution}, we provide a detailed analysis of iterated learning, followed by model ablations in Sec.~\ref{subset:ablation}. We start by describing implementation details.
\subsection{Experiment Setup}
\noindent\textbf{Training.} 
We utilize controlled experimental settings to ensure fair comparisons across models.
We train our model and all the baselines on both CC3M and CC12M datasets~\cite{sharma2018conceptual}.
For the vision agent, we adopt the default Vision Transformer (ViT-B/32) architecture~\cite{dosovitskiy2020image}, while the language agent is the same basic transformer architecture as the text encoder in CLIP~\cite{radford2021learning}. Following ~\cite{chen2023revisiting},
the codebook contains $16,384$ codes, each a $512$-dimensional vector. 
In CC3M, we set $T_{warm}, T_{distill}, T_{interact}$ to be 6000, 1000, and 5000 steps respectively.
We extended the training of the model with the final generation's parameters for additional 12k steps to ensure better convergence. We use a batch size of 1024. Detailed hyperparameter settings are available in the appendix.

\begin{figure}[t]
  \centering
  \begin{minipage}[b]{0.9\linewidth}
    \includegraphics[width=\linewidth]{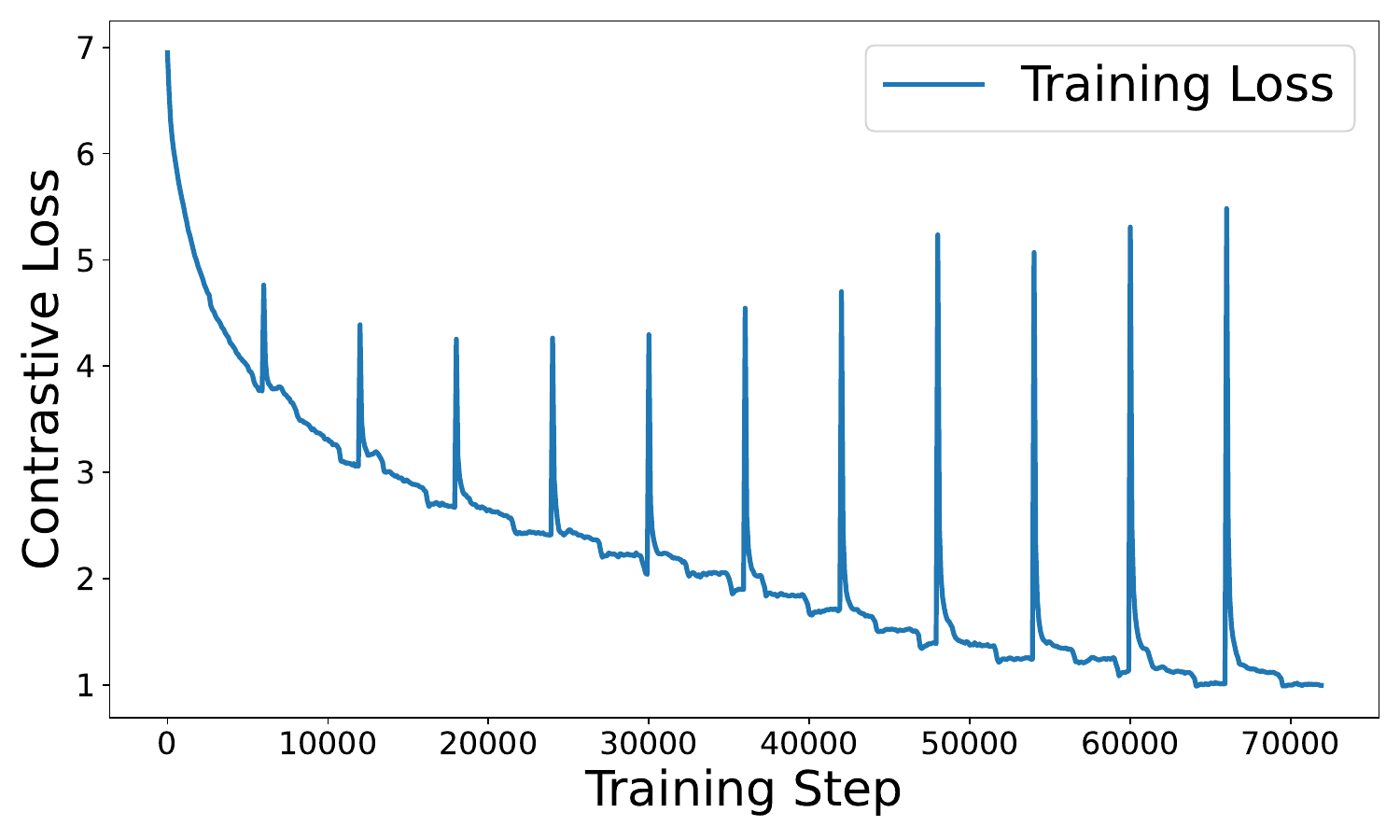}
    \caption{\textbf{Iterated learning loss curve.} Cross-modality alignment steadily improves across generations.}
    \label{fig:loss_curve}
  \end{minipage}
  \hspace{0.5cm} %
\vspace{-5pt}
\end{figure}

\vspace{5pt}
\noindent\textbf{Baseline models.}
We compare our method with standard CLIP~\cite{radford2021learning}, CLIP augmented with codebook (codebook-CLIP)~\cite{chen2023revisiting}, and CLIP enhanced through negative mining for improving compositionality (NegCLIP)~\cite{yuksekgonul2022and}. 
Hard negative mining assumes an underlying compositional structure and produces hard negatives given that structure. As such, NegCLIP serves as an unfair baseline that has additional information about how the compositionality evaluation sets were constructed.
We follow the NegCLIP design in ~\cite{yuksekgonul2022and}, with the difference that we are training from scratch. 
We create text negatives by swapping linguistic elements. We generate image negatives by maintaining a running pool of image representations, from which we extract the nearest neighbors for each batch. For a fair comparison, all models (except for NegCLIP) are trained using identical dataset and training protocols. 
\modify{ NegCLIP sees twice the amount of text data because of the hard negatives, and takes \textasciitilde 1.5x more steps to train. }

\begin{figure*}[t]
  \centering
  \begin{minipage}[b]{0.35\linewidth}
\includegraphics[width=\linewidth]{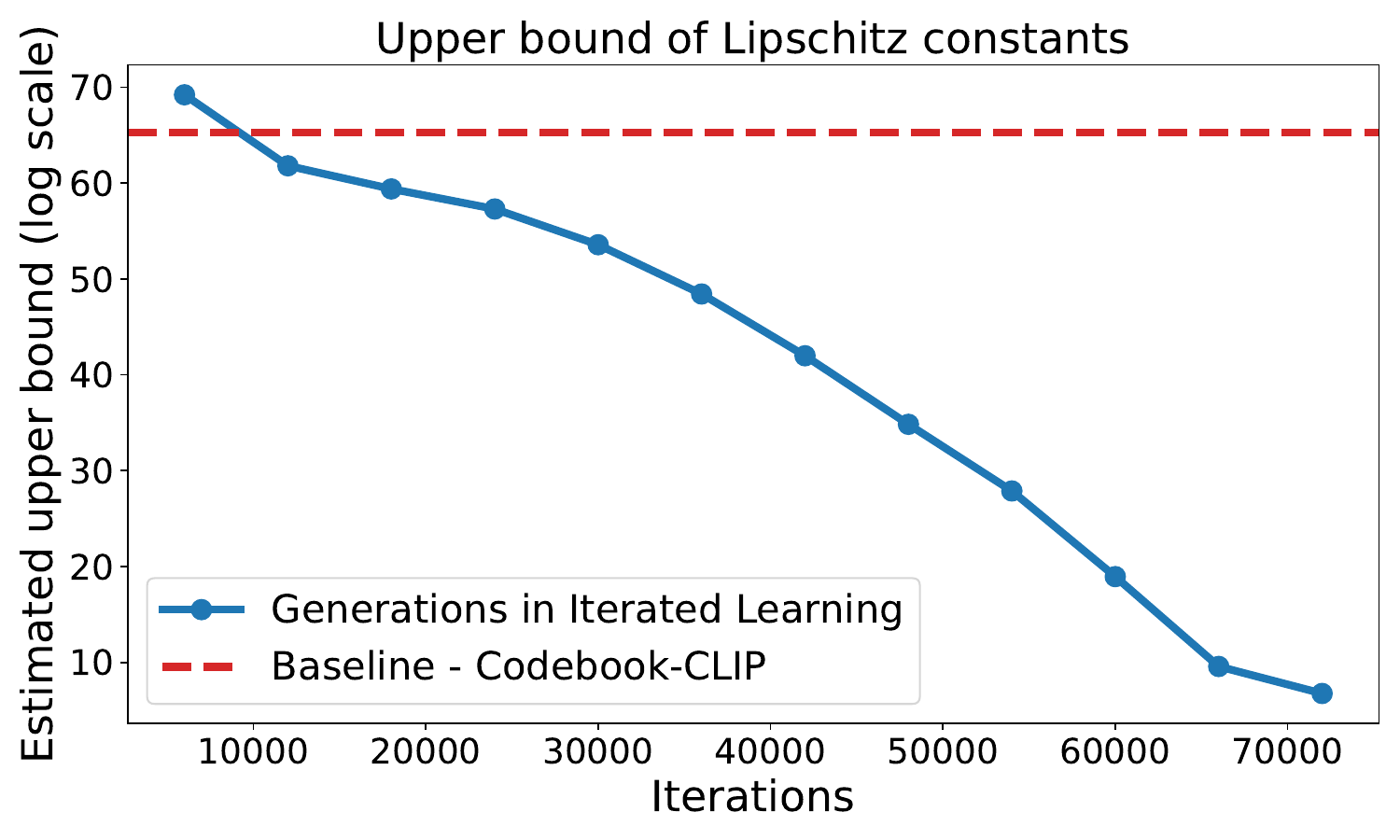}
    \caption{\textbf{Estimated Upper bound of Lipschitz Constant for  Codebook-CLIP and different generations of IL-CLIP (log scale).}}
    \label{fig:lipsconst}
  \end{minipage}
  \hspace{0.5cm} %
  \begin{minipage}[b]{0.60\linewidth}
    \includegraphics[width=\linewidth]{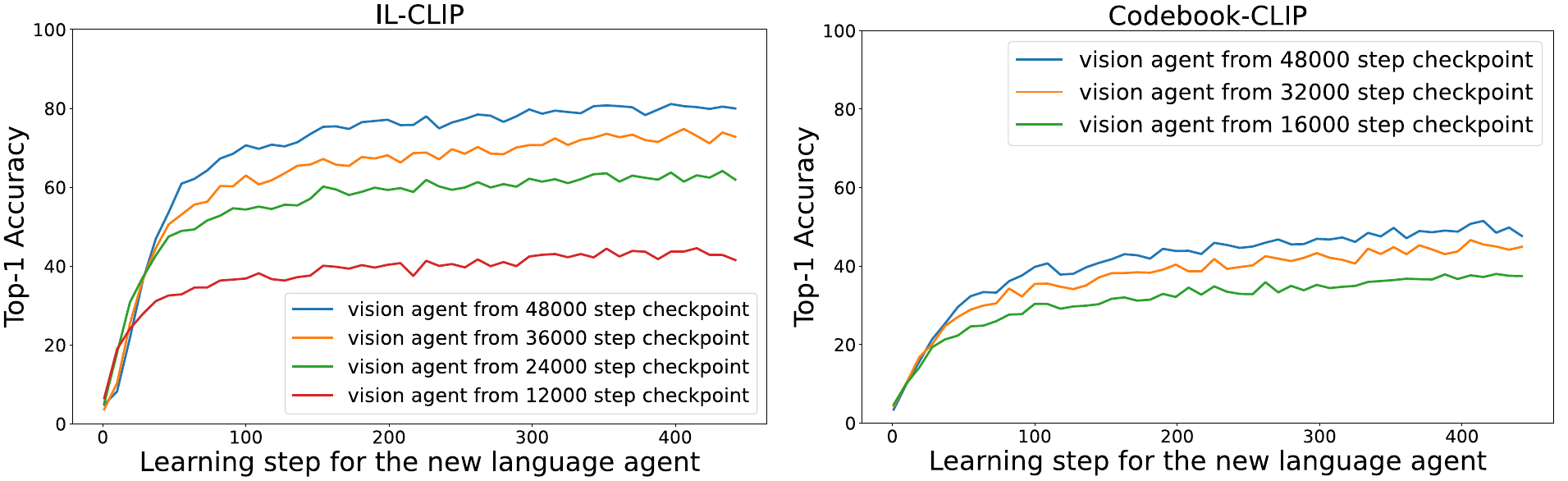}
    \caption{\textbf{Plot of in-batch image text accuracy vs. training step when a new language encoder is trained to align with fixed visual representation.} We compare between visual representation produced with iterated learning (left) and without iterated learning (right).
    }
    \label{fig:learn_from_vision}
  \end{minipage}
\vspace{-5pt}
\end{figure*}

\subsection{Iterated learning improves compositionality}
\label{subset:compositionality}

We evaluate compositionality using SugarCrepe \cite{hsieh2023sugarcrepe}, CREPE \cite{ma2022crepe}, Cola \cite{ray2023cola}, and Winoground \cite{diwan2022winoground} (Tab.~\ref{tab:comp_result}). %
These benchmarks contain image-text retrieval tasks with compositional hard-negative distractions. 
CREPE and SugarCrepe generate hard negative captions by swapping, replacing, or adding linguistic elements,  
whereas Cola and Winoground feature hand-curated hard negative images with similar visual elements but differing semantic meanings. We show examples of these data in Tab.~\ref{tab:qual}. 
Tab.~\ref{tab:comp_result} shows image-to-text retrieval accuracy for CREPE and SugarCrepe, alongside text-to-image retrieval accuracy for Cola and Winoground.

Our model outperforms all baselines in most datasets and shows notable improvement over standard CLIP. In particular, our model improves CLIP more significantly than NegCLIP, which sees text negatives close to the data in benchmarks in training time. NegCLIP achieves high scores in subsets that are close to its training objective (e.g. word swapping and negating), but fails to generalize to other hard negative types. 
Codebook-CLIP also gains performance improvement over CLIP, perhaps because the sparse codebook weight cleans supervision when facing part-of-image matches part-of-text scenarios.
So the improvement of our IL-CLIP is contributed both by the iterated learning paradigm and by the shared codebook.

\subsection{Iterated learning doesn't harm recognition}
\label{subset:recognition}

We evaluate how iterated learning affects image recognition,  following the common practice of evaluating zero-shot image classification. We report the zero-shot image-text retrieval and linear probing performance in the appendix.
We conduct the zero-shot image classification on 18 widely-used datasets (Tab.~\ref{tab:classification}), including both standard recognition datasets and datasets from the VTab benchmark \cite{zhai2019large} that measure the model's robustness. 

In line with findings from \cite{chen2023revisiting}, we also observe improvements for Codebook-CLIP over the standard CLIP model. Benefiting from the shared codebook, IL-CLIP also improves standard CLIP performance. We observe that using iterated learning on top of CLIP-codebook downgrades its performance slightly, but the difference is minimal, and IL-CLIP ranks the best in several datasets.
NegCLIP, however, performs notably worse than standard CLIP.
This is perhaps because compositionality is often viewed to be in opposition to tasks that improve with \textit{context}. Intuitively, if a model
uses \textit{context} to predict the existence of the ``road`` when
it sees a ``car``, it will increase performance on recognition
benchmarks but is not compositional. Such contextual biases
are commonplace in vision benchmarks, causing compositionality to be at odds with recognition 
Surprisingly, iterated learning renders on-par performance compared with its normal training counterparts.
Thus, we conclude that the iterated learning paradigm does not harm recognition.

\subsection{Analysis on iterated learning}
\label{subsec: evolution}
We provide a detailed analysis of iterated learning here, including evidence that IL produces easy-to-learn representations, improvement of cross-modality alignment across generations, and interpretability in the codebook.

\noindent\textbf{Iterated learning produces easy-to-learn visual representation.}
As shown in cognitive science studies~\cite{carr2017cultural, kirby2014iterated, perfors2002simulated}, compositional languages are easy-to-learn. While it is difficult to explicitly prove that the learned visual representations are compositional, we design an experiment to demonstrate they are easy-to-learn by new language agents.
In particular, given a visual agent and the codebook from a certain generation, we fix their weights and use them to train a new language agent via contrastive loss.
We target to observe how well a language agent can  ``learn'' to align its representation from different well-trained visual agent ``teachers''. 
We evaluate both our IL-CLIP (with iterated learning) and codebook-CLIP (without iterated learning). The spawned language agents in all runs are initialized using the same random weights. The results are shown in Fig.~\ref{fig:learn_from_vision}.
We find the language agents paired with vision representations developed through iterated learning achieved significantly higher matching accuracy, implying enhanced ease of learning. This is further underscored by the steeper initial slope of accuracy curves of IL-CLIP, indicating the faster learning speed for the new language agent. Thus, we conclude that IL-trained visual representation is significantly easier to learn and therefore has more chance to be compositional. Additionally, we observe from the curves of IL-CLIP that the top-1 accuracy is much higher if visual representations from later generations are used, suggesting that the property of being easy to learn progressively evolves across generations.

\noindent\textbf{Iterated learning performs smoothness regularization}
\modify{ We find that iterated learning can be seen a smoothness regularization by comparing the Lipschitz constant between our models and codebook-CLIP. While the exact Lipschitz constant for a complex model is intractable, we can estimate the upper bound of Lipschitz constant~\cite{gouk2021regularisation}. As shown in Fig.~\ref{fig:lipsconst}, the estimated upper bound of Lipschitz constant decreases as generation increases in iterated learning setting and is much smaller than the model trained with the standard scheme.}

\noindent\textbf{Cross-modality alignment steadily improves across generations.}
The contrastive loss measures the cross-modality alignment between image-text pairs. We plot the training loss for one of our IL-CLIP models (Fig.~\ref{fig:loss_curve}). Despite the big increase in loss when a new language agent is spawned, the loss still decays smoothly across generations. 
We attribute this to the representations becoming easier to learn, so 
the new language agents need fewer iterations to reach the alignment of the last generation and start to improve further.

\begin{figure}[h]
  \centering
    \includegraphics[width=0.5\textwidth]{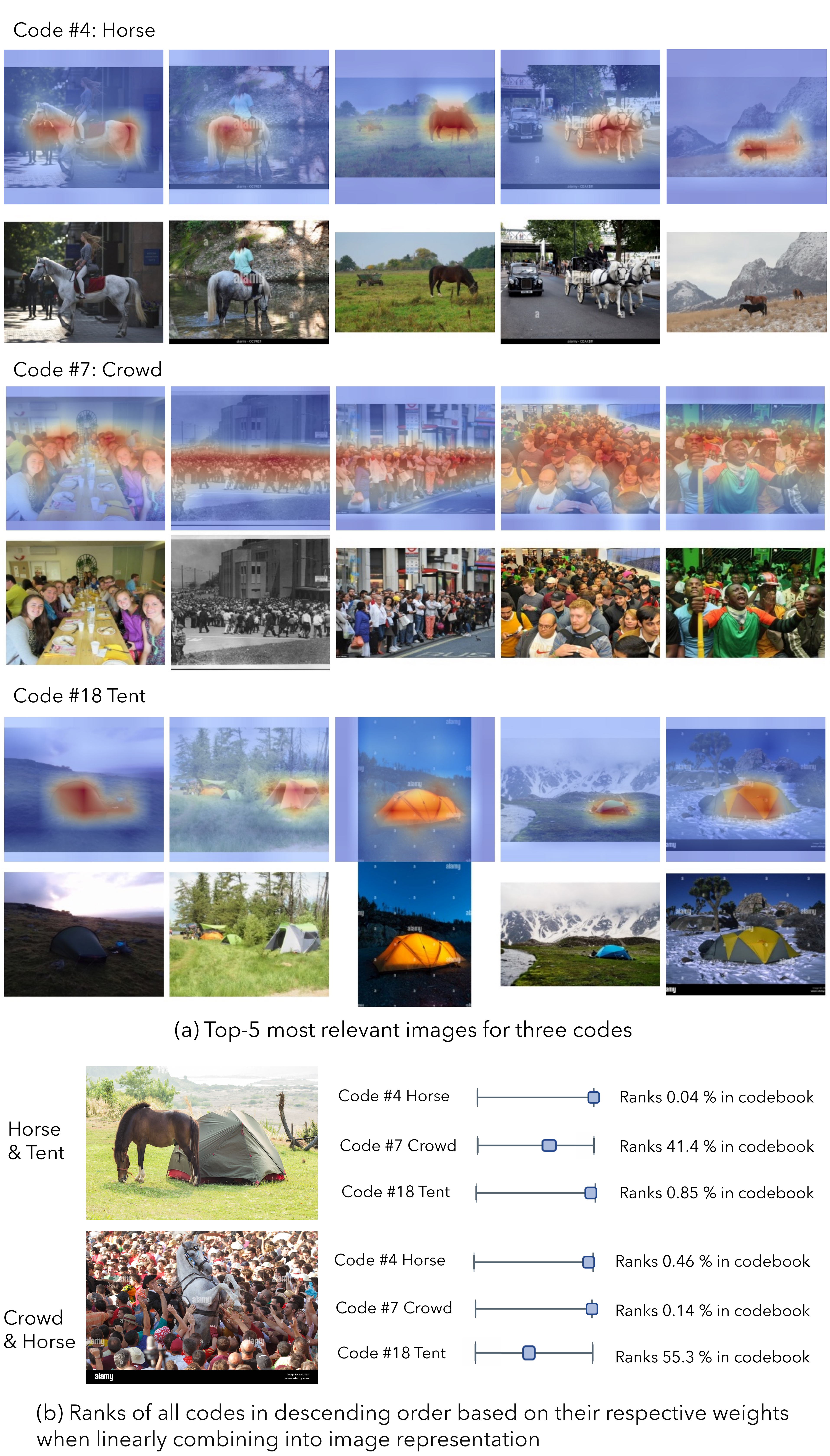}
    \caption{\textbf{Visualization of the codebook.} Most of the codes in the evolved codebook are well-grounded to specific semantic meanings, and we found some of them align with human vocabulary. We can also visualize the model's compositional reasoning by measuring how much each code contributes to the image representation.} 
    \label{fig:codebook}
    \vspace{-4mm}
\end{figure}

\noindent\textbf{The evolved codebook is (mostly) interpretable.}
We visualize the learned codebook by retrieving the top 5 most relevant images for each code (using Eq~\ref{eq:relevant}). 
We find that the codes correspond to different (somewhat) interpretable semantic concepts 
In Fig.~\ref{fig:codebook}(a), we show three examples of codes that happen to align with human vocabulary, while we show the foremost codes (sorted by index) in the appendix to ensure unbiased evidence.
After mapping the codes manually, we can reverse the process and interpret which codes are selected when viewing a new image (Fig.~\ref{fig:codebook}(b)).
For example, both the ``horse'' and ``tent'' codes are assigned a higher weight when viewing an image that contains both, indicating the model's compositional understanding. 
\modify{We find that such interpretations are harder to find in codebook-CLIP (e.g. Fig.~\ref{fig:codebook_comp}), which is shown via a user study in the appendix.} 

\begin{figure}[h]
  \centering
    \includegraphics[width=0.48\textwidth]{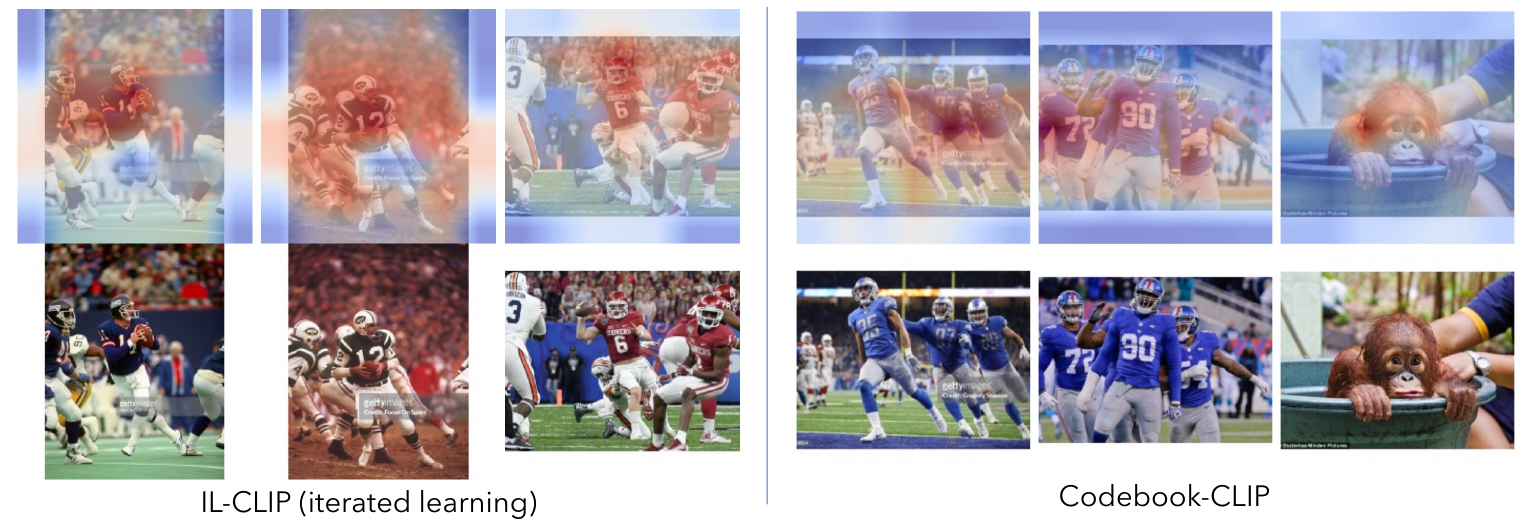}
    \caption{\textbf{Comparison of codebook interpretability.} As an example, we 
    retrieve Top-3 most relevant images for the “football player” code, and find IL-CLIP produces more consistent images. } 
    \label{fig:codebook_comp}
\end{figure}

\subsection{Ablation Study}
\label{subset:ablation}
We ablate the training duration for each generation, which agent to reset, and the choice to freeze the codebook during distillation.
All models are trained on CC3M dataset.

\noindent\textbf{Generation cycle.}  We train three models using different numbers of steps for each generation while ensuring the same total number of training steps. Tab.~\ref{tab:ablations} shows that both too few and too many steps will result in a decrease in compositionality performance, while the recognition performance is positively related to the number of steps. We hypothesize that, on one hand, the interacting agents may not be able to produce reasonably aligned representation in a very short generation cycle, and the resulting low recognition performance can negatively influence compositionality, demonstrated in \cite{hsieh2023sugarcrepe, ma2022crepe}. On the other hand, a long generation cycle enables agents to converge better in one generation, potentially leading to better recognition.
However, the reduction of generational transition frequency possibly decreases the chance to evolve more compositional representation.

\noindent\textbf{Which agent to spawn?} We experiment with resetting only language/vision agents and resetting both alternatively. Resetting only language agents renders the best performance. The alternative reset setting significantly downgraded the performance, suggesting ensuring the continuity of at least one side of agent weight is necessary for preventing the loss of recognition capacity. Interestingly, spawning language agents exhibit better performance than resetting vision agents, although the training paradigm is entirely symmetric. This is perhaps because vision agents need to learn low-level feature extractors before obtaining high-level concepts while the text is naturally abstracted by humans, therefore resetting vision agents would require more re-training efforts.

\begin{table}[t]
\centering
\resizebox{.99\linewidth}{!}{
  {\def\arraystretch{1.1}
\begin{tabular}{clccc}
  \toprule
  \textbf{Study Objectives} & \textbf{Variation} & \textbf{COMP} & \textbf{CLS}  \\
  \midrule
  \multirow{3}{*}{Generation Cycle} & 3k step & 32.1 & 23.8 \\
& 6k steps & \textbf{34.4} & 24.2  \\
  & 12k steps & 32.9 & \textbf{24.3}   \\
  \midrule
  \multirow{3}{*}{Spawn Target} & Language Agent & \textbf{34.4} & \textbf{24.2}  \\
  & Vision Agent & 33.9 & 24.0 \\
  & Alternatively & 30.7 & 21.4\\
  \midrule
  \multirow{2}{*}{Codebook Continuity} 
  & w weight fixed & \textbf{34.4} & \textbf{24.2}  \\
  & w/o weight fixed & 31.9 & 24.0 \\
  \midrule
  \multirow{2}{*}{IL w/wo Codebook} 
  & w codebook &  \textbf{34.4} & \textbf{24.2} \\
  & w/o codebook & 28.0 & 21.5 \\
  \midrule
  \multirow{2}{*}{Lipschitz Regularization} 
  & Iterated learning & \textbf{34.4} & \textbf{24.2} \\
    & L-Regularized  &  27.8 & 21.0 \\
  \bottomrule
\end{tabular}}}
\vspace{-1mm}
\caption{{\bf Ablation study}: ``COMP'' represents average scores of compositional benchmarks in Sec.~\ref{subset:compositionality}. ``CLS'' represents average scores of image classification (same  datasets as in Sec.~\ref{subset:recognition})}
\label{tab:ablations} 
\vspace{-4mm}
\end{table}

\noindent\textbf{Frozen codebook.} We study the necessity of enforcing the continuous evolution of the codebook. We train another model without fixing the codebook weight at the start of each generation. According to Table~\ref{tab:ablations}, this downgrades both the compositionality and recognition performance, since the randomly initialized weight of the newly initialized agent may contaminate the codebook.

\noindent\textbf{IL w/wo codebook.} Finally, we compare our method with/without the codebook. The results demonstrate the efficacy of using a codebook for iterated learning, since our method without the codebook underperforms its counterpart with the codebook under both compositionality and image classification evaluations.

\noindent\textbf{IL vs. Lipschitz Regularization.}
\modify{In Sec.~\ref{subsec: evolution}, we show that iterated learning performs smoothness regularization and reduces Lipschitz constant. A natural question is if Lipschitz regularization can achieve the same effect as iterated learning. We therefore trained a variant of Lipschitz-regularized CLIP that applies spectral normalization after each linear layer. As shown in Tab.~\ref{tab:ablations}, the model trained with only Lipschitz regularization barely improves performance.}

\section{Discussion}

\noindent\textbf{Conclusions.} In this paper,  we design an iterated learning algorithm that improves the compositionality in large vision-language models, inspired by cultural transmission theory in cognitive science. To achieve this, we treat vision-language contrastive learning as two agents playing the Lewis Signaling Game, and iteratively spawning new language agents by resetting weights. Our model demonstrates improvements in compositional understanding over the standard CLIP across various benchmarks, while maintaining comparable recognition capabilities. This work paves the way for future advancements in other areas requiring compositional understanding, suggesting the potential applicability of iterated learning in a broader range of tasks.

\noindent\textbf{Limitations.} Similar to the findings in cognitive science \cite{carr2017cultural}, we observe that the learning process of IL-CLIP could be unstable due to the randomness in spawning new agents. More work is needed to stabilize the learning process.

{
\small
\bibliographystyle{ieeenat_fullname}
\bibliography{main}
}

\clearpage
\setcounter{page}{1}
\appendix

\section{Implementation Detail}
\label{sec:implementation}
We state the implementation details of training and evaluating models in this section.

\noindent\textbf{Training.} We list the hyperparameters we used in pretraining in Table \ref{tab:pre-training hyperparams}. Note that our learning rate schedule during training is different from most vision-language models: we perform a linear warm-up at the start of every generation, but the long-term trend follows cosine annealing decay.
\begin{table}[h]
    \centering
    \centering
     \begin{tabular}{l|c}
        \toprule Hyperparameter & Value \\
        \midrule
        optimizer type & AdamW \\
        base learning rate & 0.0005 \\
        weight decay & 0.1 \\
        $\beta_{1}$ & 0.9 \\
         $\beta_{2}$ & 0.98 \\
        lr scheduler & Cosine Annealing \\
        warmup step & 500 for every generation \\
        image resolution & 224 \\
        max token number & 77 \\
        \bottomrule 
    \end{tabular}
    \caption{Common hyperparameters used for IL-CLIP pre-training.}
    \label{tab:pre-training hyperparams}
    \end{table}
    
\noindent\textbf{Zero-shot image classification.}
We represent each class by 
its text description. After extracting the image feature from a target image and text features for all class names, the category of the image can be predicted by choosing the class with the maximum cosine similarity score between its text feature and the image feature. We use the same multiple prompt types as in CLIP paper \cite{radford2021learning}, and the final predictions are averaged between prompts.

\section{Additional Experiments on Recognition}
\label{sec:recognition}
We supplement the recognition evaluation by doing linear probing and zero-shot image-text retrieval tasks.

\noindent\textbf{Linear probing.}
In this evaluation, we classify images by training a linear network layer on top of extracted vision features. Following CLIP \cite{radford2021learning}, We train a logistic regression classifier with L-BFGS optimizer. We set the base learning rate to be 0.05 with no weight decay. The results are shown in table \ref{tab:linear_prob}. From the fact that our model performs equally well with codebook-CLIP and much better than standard CLIP, we claim the vision representation trained by iterated learning is as powerful in recognition as the normally-trained vision representation.

\begin{table}[h]

    \centering
\scalebox{0.60}{
    \begin{tabular}{l l c c c c c c c c c c c c c c c c c c c} 
        \toprule
        \textbf{\rotatebox[origin=lb]{90}{Pretrain}}  &  
        \textbf{Method}  &  
        \rotatebox[origin=lb]{90}{\textbf{ImageNet1k}} &  
        \rotatebox[origin=lb]{90}{\textbf{CIFAR-100}} &  
        \rotatebox[origin=lb]{90}{\textbf{CIFAR-10}} &  
        \rotatebox[origin=lb]{90}{\textbf{STL-10}} &  
        \rotatebox[origin=lb]{90}{\textbf{VOC2007}} &  
        \rotatebox[origin=lb]{90}{\textbf{Caltech101}} &  
        \rotatebox[origin=lb]{90}{\textbf{Pets}} &  
        \rotatebox[origin=lb]{90}{\textbf{Flowers102}} &  
        \rotatebox[origin=lb]{90}{\textbf{Food101}} &  
        \rotatebox[origin=lb]{90}{\textbf{Mean}}
        \\ 
        
        \midrule %
        & CLIP \cite{radford2021learning} & 0.43 & 0.58 & 0.80 & 0.89 & 0.73 & 0.80 & 0.54 & 0.63 & 0.45 & 0.65  \\
        \multirow{2}{*}{\rotatebox[origin=lb]{90}{CC3M}} & {\small Codebook-CLIP} \cite{chen2023revisiting} &  0.47 & \textbf{0.62} & \textbf{0.84} & 0.90 & 0.74 & \textbf{0.82} & 0.54 & 0.64 & \textbf{0.52} & \textbf{0.68} \\
        & NegCLIP \cite{yuksekgonul2022and} & 0.39 & 0.56 & 0.79 & 0.85 & 0.72 & 0.84 & 0.46 & 0.55 & 0.40 & 0.62  \\
        & \textbf{IL-CLIP (Ours)} & \textbf{0.49} & 0.57 & 0.80 & \textbf{0.93} & \textbf{0.76} & \textbf{0.82} & \textbf{0.57} & \textbf{0.66} & 0.49 & \textbf{0.68} \\

        \midrule %
        \multirow{4}{*}{\rotatebox[origin=lb]{90}{CC12M}} & CLIP \cite{radford2021learning} & 0.60 & 0.65 & 0.85 & 0.95 & 0.76 & 0.83 & 0.72 & 0.69 & 0.64 & 0.74 \\
         &{\small Codebook-CLIP}\cite{chen2023revisiting} & \textbf{0.62} & \textbf{0.71} & \textbf{0.89} & 0.96 & \textbf{0.80} & 0.88 & 0.76 & \textbf{0.76} & \textbf{0.72} & \textbf{0.79} \\
        & NegCLIP \cite{yuksekgonul2022and} & 0.59 & 0.63 & 0.82 & 0.95 & 0.75 & 0.84 & 0.64 & 0.70 & 0.61 & 0.72 \\
        & \textbf{IL-CLIP (Ours)} & \textbf{0.62} & {0.67} & 0.85 & \textbf{0.97} & 0.78 & \textbf{0.90} & \textbf{0.79} & 0.74 & \textbf{0.72} & \textbf{0.79}  \\

        \midrule %
        \multirow{4}{*}{\rotatebox[origin=lb]{90}{DataComp}}  & CLIP \cite{radford2021learning} & 0.44 & 0.66 & 0.85 & 0.84 & 0.79 & 0.82 & 0.43 & 0.60 & 0.52 & 0.66  \\
        &  {\small Codebook-CLIP} \cite{chen2023revisiting} & \textbf{0.47} &\textbf{0.69} & \textbf{0.89} & 0.87 & 0.79 & \textbf{0.83} & \textbf{0.46} & \textbf{0.65} & \textbf{0.53} & \textbf{0.69} \\
        & NegCLIP \cite{yuksekgonul2022and} & 0.41 & 0.60 & 0.79 & 0.84 & 0.76 & 0.81 & 0.44 & 0.58 & 0.48 & 0.63 \\
        & \textbf{IL-CLIP (Ours)} & 0.45 & 0.68 & 0.87 & \textbf{0.88} & \textbf{0.80} & \textbf{0.83} & 0.45 & 0.63 & \textbf{0.53} & 0.68  \\
        \bottomrule
    \end{tabular}
    }
    \caption{
    \textbf{Evaluation on Linear probing for all model variants.}
    }
    \label{tab:linear_prob}
\end{table}

\begin{table}[h]
\centering

\scalebox{0.68}{

\setlength{\tabcolsep}{5pt}
\begin{tabular}{l l cc cc cc c} 
    \toprule
    \multirow{2.5}{*}{   \textbf{Pretrain}    } & 
   \multirow{2.5}{*}{\textbf{Method}} & 
    \multicolumn{2}{c}{\textbf{COCO}} & 
    \multicolumn{2}{c}{\textbf{Flickr8k}} &
    \multicolumn{2}{c}{\textbf{\small Flickr30k}} & 
    \multirow{2.5}{*}{\textbf{Mean}} \\
    
    \cmidrule(lr){3-4}
    \cmidrule(lr){5-6}
    \cmidrule(lr){7-8}

   & & {\small IR} & {\small TR } & {\small IR } & {\small TR }& {\small IR }& {\small TR } \\ 
    \midrule

    & CLIP~\cite{radford2021learning} & 0.23 & 0.28 & 0.41 & 0.50 & 0.39 & 0.48  & 0.38    \\
    \multirow{2}{*}{CC3M} & Codebook-CLIP~\cite{chen2023revisiting} & \textbf{0.28} & \textbf{0.35} & \textbf{0.47} & \textbf{0.57} & \textbf{0.46} & \textbf{0.57}      & \textbf{0.44}   \\
    & NegCLIP ~\cite{yuksekgonul2022and} & 0.19 & 0.23 & 0.35 & 0.42 & 0.31 & 0.38  & 0.31\\
    & \textbf{IL-CLIP (Ours)} & \textbf{0.28} & 0.32 & 0.46 & \textbf{0.57} & 0.42 & 0.51   & 0.42       \\
    \midrule
    & CLIP~\cite{radford2021learning} & 0.39 & 0.53 & 0.60 & 0.75 & 0.60 & 0.73 & 0.60 \\
    \multirow{2}{*}{CC12M} & Codebook-CLIP~\cite{chen2023revisiting} & \textbf{0.45} & \textbf{0.59} & \textbf{0.65} & \textbf{0.81} & \textbf{0.65} & \textbf{0.81} & \textbf{0.66}\\
    & NegCLIP ~\cite{yuksekgonul2022and} & 0.36 & 0.48 & 0.57 & 0.69 & 0.56 & 0.68 & 0.56\\
    & \textbf{IL-CLIP (Ours)} & 0.44 & 0.56 & 0.63 & 0.77 & 0.64 & 0.76 & 0.63\\
    \midrule
    & CLIP~\cite{radford2021learning} & 0.16 & 0.21 & 0.24 & 0.31 & 0.23 & 0.32 &  0.24 \\
    \multirow{2}{*}{DataComp} & Codebook-CLIP~\cite{chen2023revisiting} & \textbf{0.20} & \textbf{0.25} & \textbf{0.26} & \textbf{0.36} & \textbf{0.26} & \textbf{0.35} & \textbf{0.28}\\
    & NegCLIP ~\cite{yuksekgonul2022and} & 0.13 & 0.16 & 0.23 & 0.29 & 0.19 & 0.28 & 0.21\\
    & \textbf{IL-CLIP (Ours)} & 0.18 & 0.22 & \textbf{0.26} & 0.31 & 0.24 & 0.33 & 0.26 \\
    \bottomrule
\end{tabular}
}
\vspace{-2mm}
\caption{\textbf{Zero-shot image/text retrieval.} We report retrieval R@5 scores for in three most commonly used retrieval datasets. \textit{IR} stands for image retrieval, \textit{TR} stands for text retrieval.}\vspace{-2mm} %
\label{tab:reterival}
\vspace{-1mm}
\end{table}
\setlength{\tabcolsep}{6pt}

\begin{table*}[t]
\footnotesize
  \setlength{\tabcolsep}{1.8pt}
  \setlength{\extrarowheight}{0pt}
  \renewcommand{\arraystretch}{0.0}
  \newcolumntype{L}{>{\raggedright\arraybackslash}X}
  \centering
\begin{tabularx}{\linewidth}{p{0.3cm}p{0.3cm}p{0.3cm}p{0.1cm}Lp{0.15cm}Lp{0.15cm}Lp{0.1cm}L}
  \toprule[1pt]
 && && \multicolumn{5}{c}{\bf IL-CLIP wins over CLIP} && \multicolumn{1}{c}{\bf CLIP wins over IL-CLIP} \\
  \cmidrule[0.5pt]{5-9} \cmidrule[0.5pt]{11-11}
\multirow{15}{*}{\rotatebox{90}{\hspace*{-0.7cm}\textbf{Image to Text Retrieval}}} && \multirow{10}{*}{\rotatebox{60}{\hspace*{-0.5cm}Query}}
 && \vspace{2pt}\centerline{\includegraphics[width=2.5cm,height=1.5cm]{pictures/1-1.jpg}}
 && \vspace{2pt}\centerline{\includegraphics[width=2.5cm,height=1.5cm]{pictures/1-2.jpg}}
 && \vspace{2pt}\centerline{\includegraphics[width=2.5cm,height=1.5cm]{pictures/1-3.jpg}}
 && \vspace{2pt}\centerline{\includegraphics[width=2.5cm,height=1.5cm]{pictures/1-4.jpg}}
 \\
 && \vspace{-9pt}\rotatebox{60}{\hspace*{-0.5cm}Positive}
 && \vspace{-2pt}\textit{Several \textcolor{blue}{square} pizzas are sitting on \textcolor{blue}{round} plates.}
 && \vspace{-2pt}\textit{A vase with flowers \textcolor{blue}{on} a display near a wall}
 && \vspace{-2pt}\textit{\textcolor{blue}{Two} airplanes flying in the sky above \textcolor{blue}{a black} bridge.}
 && \vspace{-2pt}\textit{ A duck floating in the water near a bunch of grass and rocks.}
 \\ && \vspace{-3pt}\rotatebox{60}{\hspace*{-0.5cm}Negative}\vspace{0pt}
 && \vspace{6pt}\textit{Several \textcolor{blue}{round} pizzas are sitting on \textcolor{blue}{square} plates.}\vspace{0pt}
 && \vspace{6pt}\textit{A vase with flowers \textcolor{blue}{next} to a display near a wall.}\vspace{0pt}
 && \vspace{6pt}\textit{\textcolor{blue}{A black} airplane flying in the sky above \textcolor{blue}{two} bridges.}\vspace{0pt}
 && \vspace{6pt}\textit{A duck floating in the water near a bunch of \textcolor{blue}{flowers}, grass, and rocks.}\vspace{0pt}
 \\ 
\midrule

\vspace{20pt}\multirow{20}{*}{\rotatebox{90}{\hspace*{-0.5cm}\textbf{Text to Image Retrieval}}}
 && \vspace{2pt}\rotatebox{60}{\hspace*{-0.5cm}Query}\vspace{3pt}
 && \vspace{2pt}\textit{A young person kisses an old person.}\vspace{3pt}
 && \vspace{2pt}\textit{There are more snowboarders than skiers.}\vspace{3pt}
 && \vspace{2pt}\textit{A person is in the water and close to the sand.}\vspace{3pt}
 && \vspace{2pt}\textit{The child is throwing the adult the ball.}\vspace{3pt}
 \\ &&  \vspace{8pt}\multirow{50}{*}{\rotatebox{60}{\hspace*{-0.5cm}Positive}}
 && \vspace{0pt}\centerline{\includegraphics[width=2.5cm,height=1.5cm]{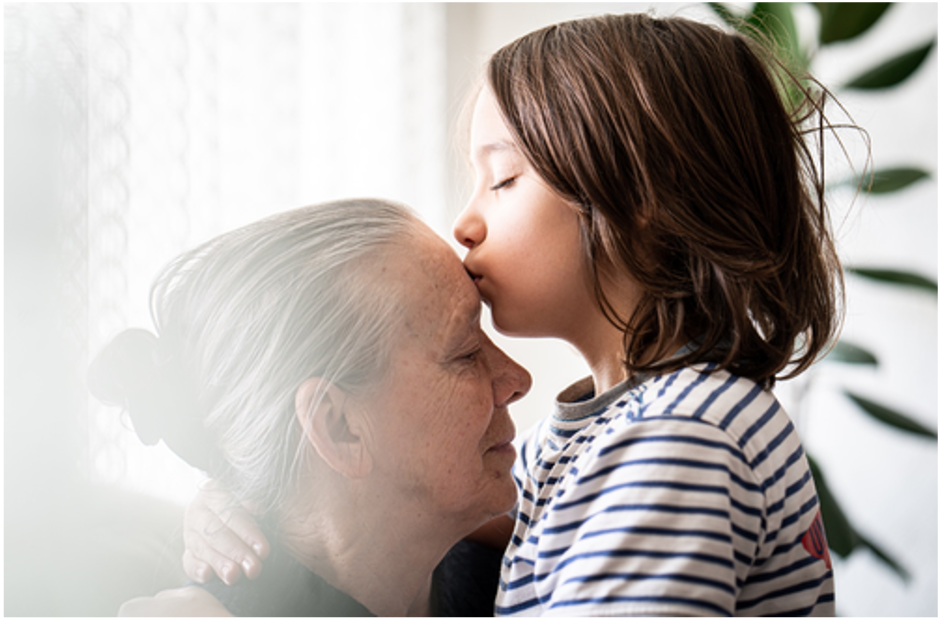}}
 && \vspace{0pt}\centerline{\includegraphics[width=2.5cm,height=1.5cm]{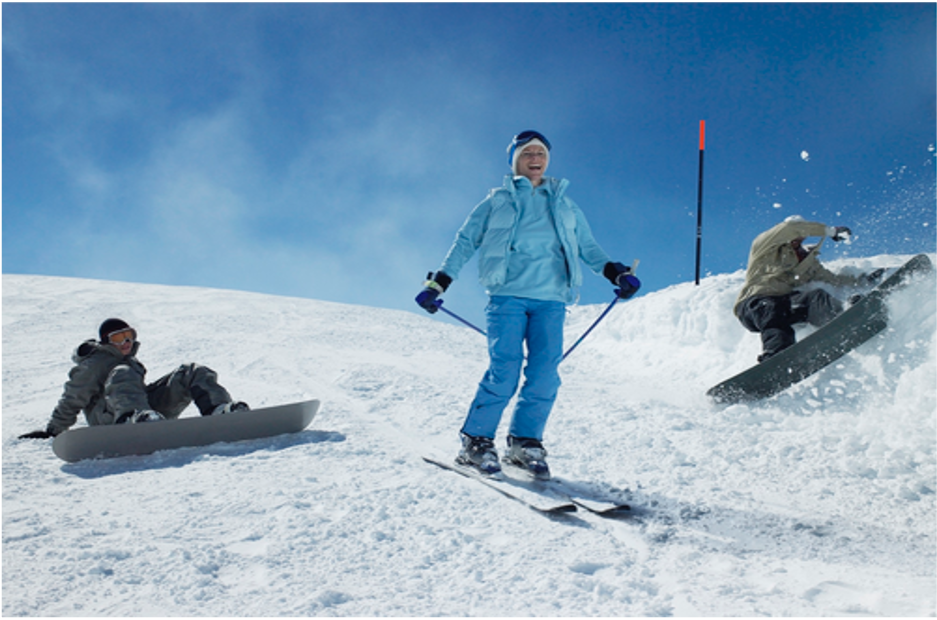}}
 && \vspace{0pt}\centerline{\includegraphics[width=2.5cm,height=1.5cm]{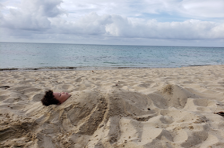}}
 && \vspace{0pt}\centerline{\includegraphics[width=2.5cm,height=1.5cm]{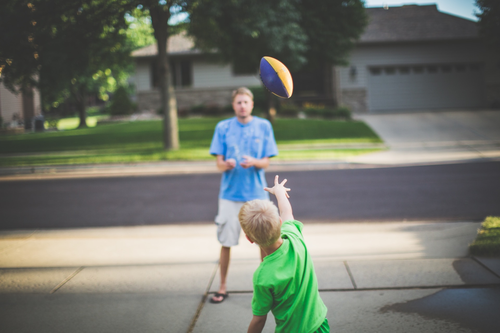}}
  \\ && \vspace{5pt}\multirow{40}{*}{\rotatebox{60}{\hspace*{-0.5cm}Negative}}\vspace{-30pt}
 && \vspace{-4pt}\centerline{\includegraphics[width=2.5cm,height=1.5cm]{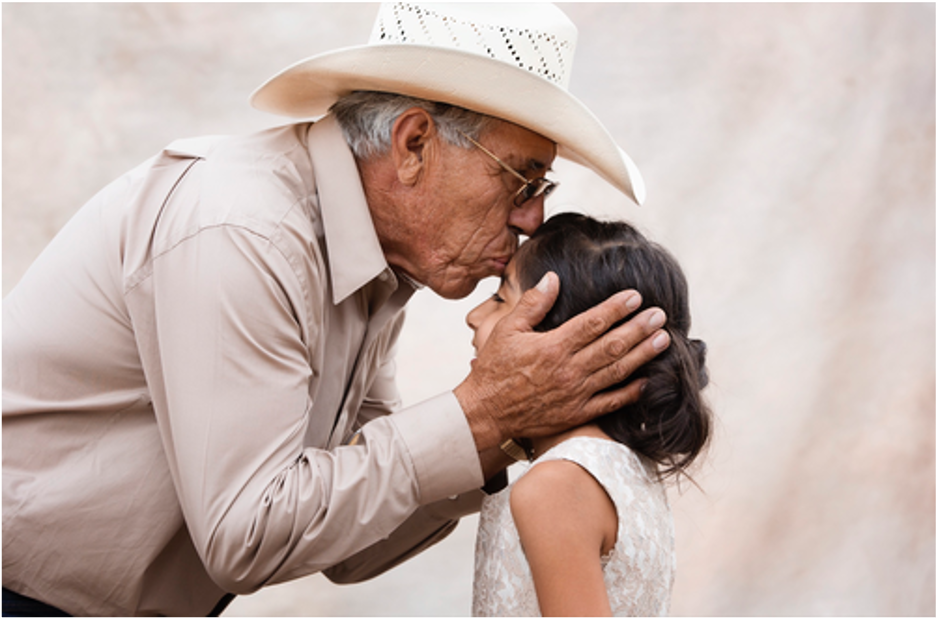}}\vspace{-30pt}
 && \vspace{-4pt}\centerline{\includegraphics[width=2.5cm,height=1.5cm]{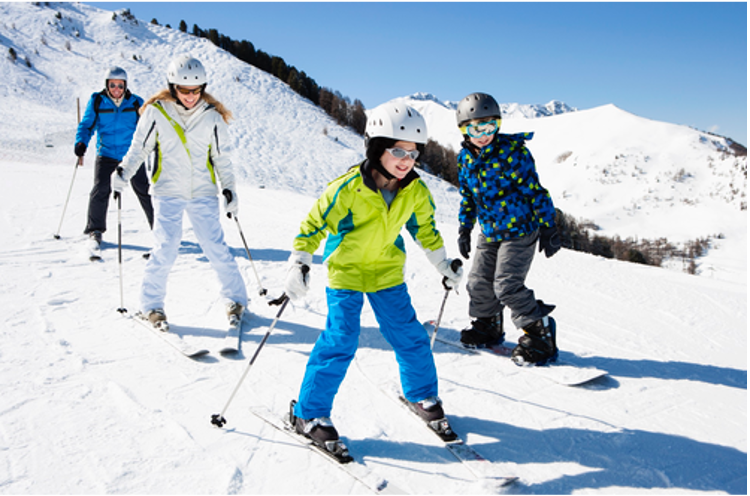}}\vspace{-30pt}
 && \vspace{-4pt}\centerline{\includegraphics[width=2.5cm,height=1.5cm]{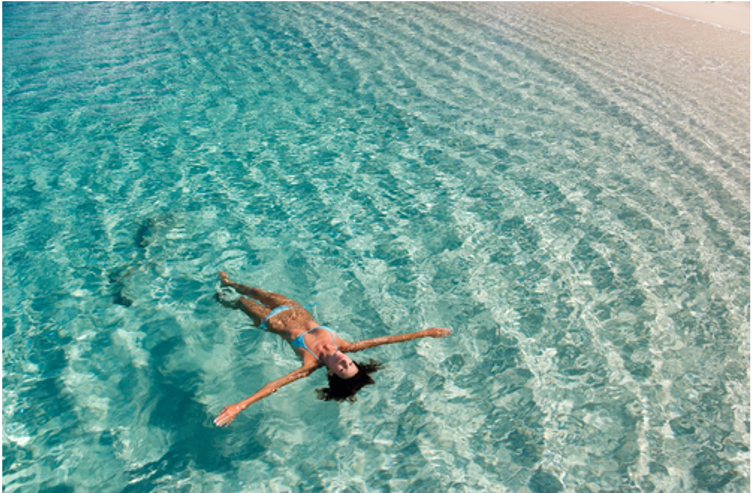}}\vspace{-30pt}
 && \vspace{-4pt}\centerline{\includegraphics[width=2.5cm,height=1.5cm]{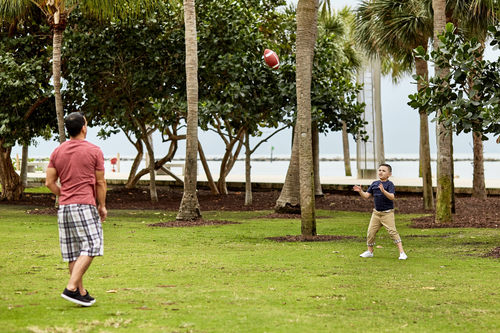}}\vspace{-30pt}
 \\
 \bottomrule
  \end{tabularx}
  \caption{
    \textbf{Sampled test cases in compositionality benchmarks and performance comparison.} We found our model exhibits better compositional understanding than standard CLIP in distinguishing compositional hard negatives.
  }\label{tab:qual}
\end{table*}

\noindent\textbf{Zero-shot image-text retrieval.}
We evaluate all models' zero-shot retrieval performance on the test set of three standard benchmarks: MS-COCO \cite{coco}, Flickr8k \cite{young2014image} and Flickr30k \cite{young2014image}. 
The performance is shown in table \ref{tab:reterival}.
While iterated
learning slightly downgrades the performance of Codebook-CLIP, it still maintains a lead over the standard CLIP model. The performance drop is potentially due to the fact that the text representation is under-trained under the iterated learning framework since the language agent is dynamically replaced.

\section{Pretraining on DataComp dataset}
We also pretrain our models on DataComp-10M dataset to ensure our finding is not specific to any pretraining dataset. 
We report the detailed compositionality and image classification accuracy in table \ref{tab:supp_comp_result} and \ref{tab:supp_classification} respectively. Their linear probing and image-text retrieval performance are shown along with other variants of models in table  \ref{tab:linear_prob} and \ref{tab:reterival}.
The noisiness of unfiltered DataComp-10M turns out to influence all models' performance, but the IL-CLIP is still the best model in compositionality and comparable to Codebook-CLIP in recognition, which is consistent with the findings in the main paper.
\section{Additional Ablation: Iterated Learning with Hard Negative Mining}
Our proposed Iterated learning algorithm augments the CLIP training procedure, while NegCLIP augments the CLIP training objective. In principle, these two approach can work together and potentially result in a better model. 
As an additional ablation, we study a variant of the CLIP model that uses both iterated learning and hard negative mining during training. We train it on the CC3M dataset.
From the results in Table \ref{tab:IL-NegCLIP}, we observe the combination of iterated learning and negative mining yields a model with the best compositionality performance, but leads to a slight performance drop for recognition.

\begin{table}[h]
\centering
\resizebox{.99\linewidth}{!}{
  {\def\arraystretch{1.1}
  \centering
\begin{tabular}{clcccccc}
  \toprule
   \textbf{Models} & \textbf{compositionality} & \textbf{classification} & \textbf{probing} & \textbf{retreival} \\
  \midrule
CLIP  & 0.28 & 0.22 & 0.65 & 0.38 \\
NegCLIP & 0.32 & 0.21 & 0.62 & 0.31 \\
IL-CLIP  & 0.34 &\textbf{0.24} & \textbf{0.68} & \textbf{0.42} \\
\colorbox{tablegreen}{IL-NegCLIP}  & \colorbox{tablegreen}{\textbf{0.35}} & \colorbox{tablegreen}{\textbf{0.24}} & \colorbox{tablegreen}{0.67} & \colorbox{tablegreen}{0.40} \\
  \bottomrule
\end{tabular}}}
\vspace{-1mm}
\caption{{\bf Iterated learning with hard negative mining. } \colorbox{tablegreen}{Color notations:} The performance of the target model that combines negative mining and iterated learning.}
\label{tab:IL-NegCLIP} 
\vspace{-4mm}
\end{table}

\begin{table*}[h]
\centering

\scalebox{0.89}{

\setlength{\tabcolsep}{5pt}
\begin{tabular}{l l  ccccc ccc cc c c c} 
    \toprule
    \multirow{2.5}{*}{\textbf{Dataset}} & 
   \multirow{2.5}{*}{\textbf{Method}} & 
    \multicolumn{2}{c}{\textbf{CREPE-systematicity}} & 
    \multicolumn{3}{c}{\textbf{CREPE-productivity}} & 
    \multicolumn{3}{c}{\textbf{SugarCrepe}} &
    \textbf{Cola} &
    \multicolumn{1}{c}{\textbf{\small Winoground}} & 
    \multirow{2.5}{*}{\textbf{\small Mean}} \\
    \cmidrule(lr){3-4}
    \cmidrule(lr){5-7}
    \cmidrule(lr){8-10}
    \cmidrule(lr){11-11}
    \cmidrule(lr){12-12}

   & & {\small atom} & {\small compound} & {\small replace} & {\small swap}& {\small negate}& {\small add} & {\small replace} & {\small swap} &  {\small Txt2Img} & {\small Txt2Img} \\ 
    \midrule

    \multirow{4}{*}{DataComp} & CLIP~\cite{radford2021learning} & 0.33 & 0.36 & 0.11 & 0.20 & 0.10 & 0.63 & 0.62 & 0.57 & \textbf{0.21} & 0.10 & 0.32 \\
    & Codebook-CLIP~\cite{chen2023revisiting} & \textbf{0.34} & 0.37 & 0.12 & 0.21 & 0.09 & 0.64 & 0.64 & 0.59 & 0.20 & 0.07 & 0.33 \\
    & NegCLIP ~\cite{yuksekgonul2022and} & 0.32 & 0.36 & 0.11 & \textbf{0.24} & \textbf{0.12} & 0.62 & 0.63 & \textbf{0.64} & 0.19 & 0.11 & 0.33 \\
    & \textbf{IL-CLIP (Ours)} & \textbf{0.34} & \textbf{0.40} & \textbf{0.14} & 0.23 & 0.09 & \textbf{0.66} & \textbf{0.66} & 0.62 & 0.18 & \textbf{0.14} & \textbf{0.35} \\
    
    \bottomrule
\end{tabular}
}
\vspace{-2mm}
\caption{\textbf{Evaluation on compositionality benchmarks for models pretrained on DataComp-10M.}
 }\vspace{-2mm} %
\label{tab:supp_comp_result}
\end{table*}
\setlength{\tabcolsep}{6pt}

\begin{table*}[h]
    
    \centering
\scalebox{0.71}{
    \begin{tabular}{l l c c c c c c c c c c c c c c c c c c c} 
        \toprule
        \textbf{Dataset}  &  
        \textbf{Method}  &  
        \rotatebox[origin=lb]{90}{\textbf{ImageNet1k}} &  
        \rotatebox[origin=lb]{90}{\textbf{CIFAR-100}} &  
        \rotatebox[origin=lb]{90}{\textbf{CIFAR-10}} &  
        \rotatebox[origin=lb]{90}{\textbf{STL-10}} &  
        \rotatebox[origin=lb]{90}{\textbf{VOC2007}} &  
        \rotatebox[origin=lb]{90}{\textbf{Caltech101}} &  
        \rotatebox[origin=lb]{90}{\textbf{SUN397}} &  
        \rotatebox[origin=lb]{90}{\textbf{Pets}} &  
        \rotatebox[origin=lb]{90}{\textbf{Flowers102}} &  
        \rotatebox[origin=lb]{90}{\textbf{Food101}} &  
        \rotatebox[origin=lb]{90}{\textbf{ObjectNet}} &  
        \rotatebox[origin=lb]{90}{\textbf{CLEVR}} &  
        \rotatebox[origin=lb]{90}{\textbf{Smallnorb}} &  
        \rotatebox[origin=lb]{90}{\textbf{Resisc45}} & 
        \rotatebox[origin=lb]{90}{\textbf{DMLAB}} & 
        \rotatebox[origin=lb]{90}{\textbf{ImageNet-A}} & 
        \rotatebox[origin=lb]{90}{\textbf{ImageNet-R}} & 
        \rotatebox[origin=lb]{90}{\textbf{IN-sketch}} & 
        \rotatebox[origin=lb]{90}{\textbf{Mean}}
        \\ 
        \midrule %
        & CLIP \cite{radford2021learning} & 0.14 & 0.31 & 0.72 & 0.72 & 0.32 & 0.62 & 0.22 & 0.12 & 0.05 & 0.16 & 0.15 & 0.12 & 0.06 & 0.12 & 0.13 & \textbf{0.02} & 0.17 & 0.09 & 0.24 \\
        \multirow{2}{*}{DataComp} & Codebook-CLIP \cite{chen2023revisiting} & \textbf{0.15} & \textbf{0.36} & \textbf{0.76} & 0.72 & 0.38 & \textbf{0.68} & \textbf{0.25} & \textbf{0.14} & \textbf{0.07} & \textbf{0.18} & 0.18 & \textbf{0.14} & 0.06 & 0.14 & 0.13 & \textbf{0.02} & \textbf{0.19} & \textbf{0.10} & \textbf{0.26} \\
        & NegCLIP \cite{yuksekgonul2022and} & 0.12 & 0.28 & 0.67 & 0.69 & 0.32 & 0.59 & 0.21 & 0.11 & 0.04 & 0.14 & 0.16 & 0.12 & 0.07 & 0.12 & 0.14 & \textbf{0.02} & 0.13 & 0.07 & 0.22 \\
        & \textbf{IL-CLIP (Ours)} & 0.14 & 0.33 & 0.74 & \textbf{0.74} & \textbf{0.42} & 0.65 & 0.24 & 0.11 & 0.06 & 0.16 & \textbf{0.19} & 0.13 & \textbf{0.08} & \textbf{0.15} & \textbf{0.15} & \textbf{0.02} & 0.16 & 0.09 & \textbf{0.26} \\
        \bottomrule
    \end{tabular}
    }
    \caption{\textbf{Evaluation of zero-shot image classification with models pretrained on DataComp-10M.}  
    }
    \label{tab:supp_classification}
\end{table*}

\section{A user study: Comparing Codebook Interpretability}
\label{appendix: user_study}
To compare the interpretability of the trained codebook between IL-CLIP and normal Codebook-CLIP, we conduct a user study where participant annotates whether randomly picked codes have semantically grounded meanings. Across \textit{50} binary decisions on whether specific codes have a semantic meaning, our \textit{10} users annotated \textit{44} codes (in average) to be interpretable in IL-CLIP versus only \textit{39} for codebook-CLIP.

\section{Unbiased Visualization for Trained Codebook (Sorted by Index)}
\begin{figure*}[t]
  \centering
    \includegraphics[width=\textwidth]{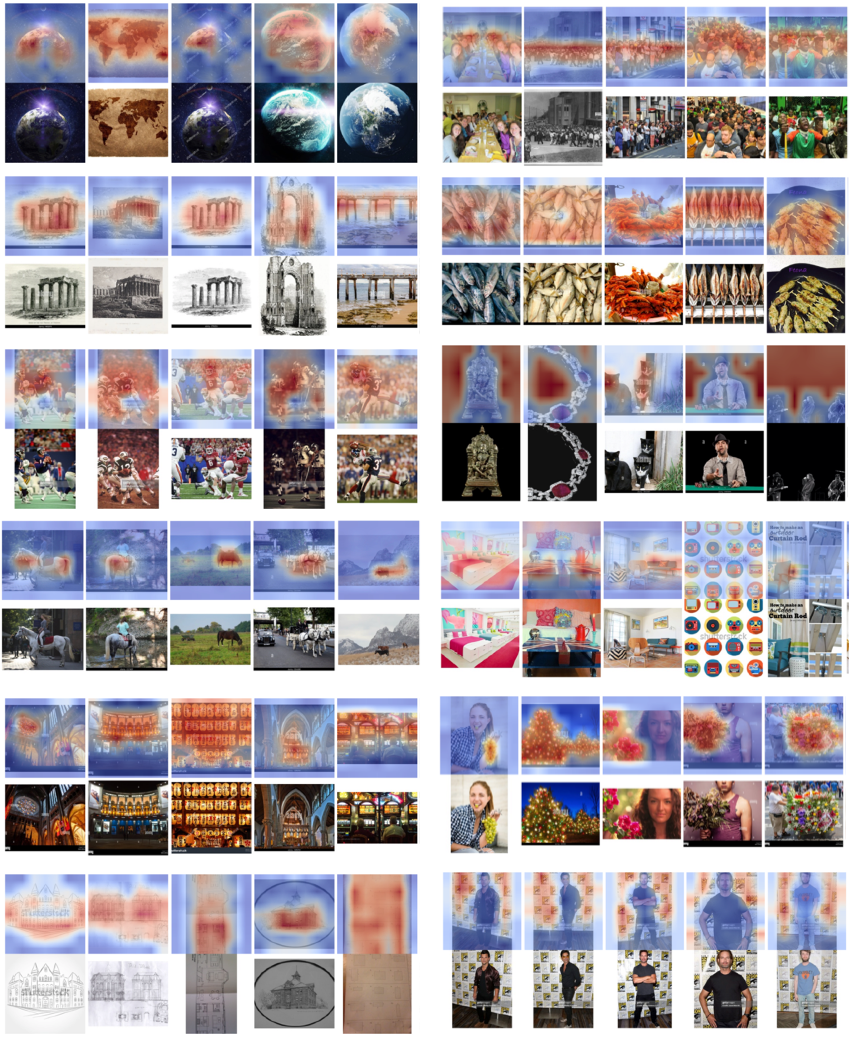}
    \caption{\textbf{Codebook visualization:} code \#1 - \#11}
    \label{fig:code1}
\end{figure*}

\begin{figure*}[t]
  \centering
    \includegraphics[width=\textwidth]{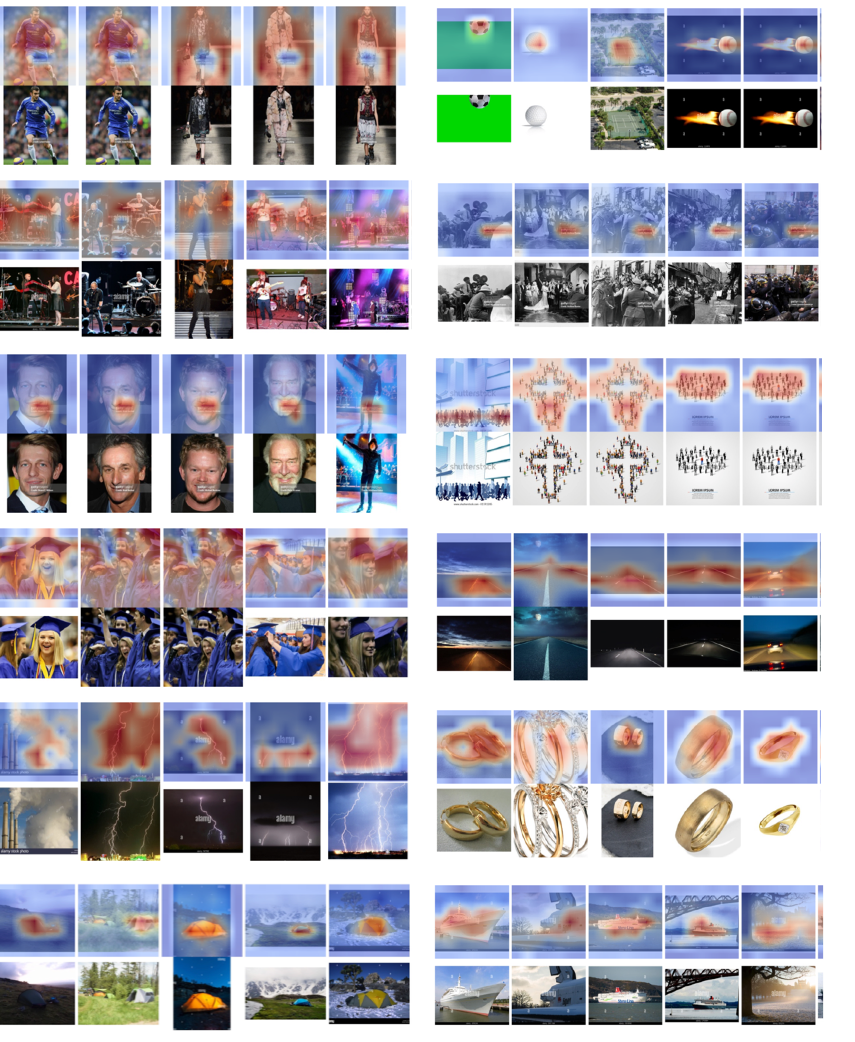}
    \caption{\textbf{Codebook visualization:} code \#12 - \#23}
    \label{fig:code2}
\end{figure*}

To unbiasedly show the performance of our trained codebook, we present a visualization of the foremost codes, organized in ascending order by their index in Fig.~\ref{fig:code1} -~\ref{fig:code2}. We find most of the codes achieve good semantic groundings, and some of them are interpretable.

\section{Qualitative Result of the Models' Performance in Compositional Understanding}
In Table~\ref{tab:qual}, we show some qualitative results, including both image-to-text and text-to-image examples. Due to enhanced compositional understanding, we observe our model does better in relationship understanding and counting.

\newpage

\end{document}